%% file: arxiv.tex
\documentclass{article}
\usepackage[letterpaper,margin=1.0in]{geometry}
\usepackage{hyperref}
\usepackage{enumitem}
\usepackage[utf8]{inputenc} %
\usepackage[T1]{fontenc}    %
\usepackage{hyperref}       %
\usepackage{url}            %
\usepackage{booktabs}       %
\usepackage{amsfonts}       %
\usepackage{nicefrac}       %
\usepackage{microtype}      %
\usepackage{graphicx}       %
\usepackage[numbers,sort]{natbib}
\usepackage{multirow}
\usepackage{bbding}
\graphicspath{{media/}}     %

\usepackage{thmtools}
\usepackage{thm-restate}

\usepackage{hyperref}
\usepackage{amsmath}
\usepackage{amsthm}
\usepackage{amssymb}
\usepackage{url}
\usepackage{cleveref}
\usepackage{enumitem}

\newtheorem{remark}{Remark}
\usepackage{enumitem}
\usepackage{graphicx}
\usepackage{caption}  %
\usepackage{subcaption}  %
\usepackage{adjustbox}  %

\crefname{figure}{Figure}{Figures}
\Crefname{figure}{Figure}{Figures}

\crefname{table}{Table}{Tables}
\Crefname{table}{Table}{Tables}

\crefname{section}{Section}{Sections}
\Crefname{section}{Section}{Sections}

\crefname{equation}{Equation}{Equations}
\Crefname{equation}{Equation}{Equations}

\usepackage{xcolor}
\usepackage[most]{tcolorbox}
\newif\ifcomments
\commentsfalse

\definecolor{huaqing}{RGB}{0,110,80}

\newif\ificml
\icmlfalse
\include{macros.tex}

\title{Configuration-to-Performance Scaling Law with Neural Ansatz}

\author{
  Huaqing Zhang \\
  IIIS, Tsinghua University \\
  \texttt{zhanghq22@mails.tsinghua.edu.cn}
  \and
  Kaiyue Wen \\
  Stanford University \\
  \texttt{kaiyuew@stanford.edu}
  \and
  Tengyu Ma  \\
  Stanford University \\
  \texttt{tengyuma@stanford.edu}
}

\date{}  %

\begin{document}
\maketitle
\begin{abstract}
Researchers build scaling laws to forecast the training performance of expensive large-scale runs with larger model size $N$ and data size $D$. These laws assume that other training hyperparameters are optimally chosen, which can require significant effort and, in some cases, be impossible due to external hardware constraints.   
To improve predictability across a broader set of hyperparameters and enable simpler tuning at scale, we propose learning a  \textit{Configuration-to-Performance Scaling Law} (CPL): a mapping from the \textit{full training configuration} to training performance. Because no simple functional form can express this mapping, we parameterize it with a large language model (LLM), and fit it with diverse open-source pretraining logs across multiple sources, yielding a \textit{Neural} Configuration-to-Performance Scaling Law (\predictor).
\predictor\ accurately predicts how training configurations influence the final pretraining loss, achieving 20-40\% lower prediction error than the configuration-agnostic Chinchilla law and generalizing to runs using up to 10$\times$ more compute than any run in the training set.
It further supports joint tuning of multiple hyperparameters with performance comparable to hyperparameter scaling law  baselines.  
Finally, \predictor\ naturally and effectively extends to richer prediction targets such as loss-curve prediction.
\footnote{Our code is available at \href{https://github.com/zhqwqwq/Configuration-to-Performance-Scaling-Law}{https://github.com/zhqwqwq/Configuration-to-Performance-Scaling-Law}.}
\end{abstract}

\input{sections/intro}

\input{sections/preliminaries}
\input{sections/methodology}

\input{sections/experiments}

\input{sections/related_work}
\input{sections/discussion}

\section*{Acknowledgement}

The authors thank Zihan Qiu, Luke Bailey, Neil Band, Caroline Choi, Arvind Mahankali, and Thomas Chen for valuable discussions and feedback.
\newpage 
\bibliography{reference}
\bibliographystyle{icml2026}

\newpage
\appendix
\input{appendix/details}
\newpage
\textbf{}
\newpage
\input{appendix/additionalresults}

\end{document}

%% file: macros.tex
\newcommand{\conf}{$\mathrm{C}$ }
\newcommand{\metr}{$\mathrm{P}$ }
\newcommand{\confnospace}{$\mathrm{C}$}
\newcommand{\metrnospace}{$\mathrm{P}$}

\newcommand{\predictors}{NCPL's}
\newcommand{\Predictor}{NCPL}
\newcommand{\predictor}{NCPL}
\newcommand{\predictornospace}{NCPL}
\newcommand{\chin}{\ell_{\mathrm{chinchilla}}}
\newcommand{\hatchin}{\hat\ell_{\mathrm{chinchilla}}}

\newcommand{\trainset}{\mathcal{D}_{\mathrm{train}}}

%% file: sections/intro.tex
\section{Introduction}

As pretraining large language models is extremely costly \citep{deepseekai2025deepseekv3technicalreport,kimiteam2025kimik2openagentic}, it is critical to have predictability of performance before executing the training run with the biggest models.  People use training runs of smaller models to build a scaling law---oftentimes a power law---that maps the number of model parameters $N$ and the amount of data (or tokens) $D$ to the predicted pretraining loss~\citep{kaplan2020openaiscaling,hoffmann2022chinchilla}. 
With an accurate scaling law,  researchers predict the pretraining performance for $N$ and $D$ that are larger than those that have been experimented with, and can decide the optimal choice of $N$ and $D$ for a given target compute (which depends on $ND$). Recent work also extends it to include a few other training hyperparameters, such as the learning rate, as inputs or outputs to facilitate choosing these hyperparameters at scale \cite{luo2025multipower,tissue2024scalinglawlearningrate,xie2024optimization, deepseekai2024deepseekscaling, porian2024porianlaw, li2025steplaw}.

\begin{figure*}[t]
    \centering
    \includegraphics[width=\linewidth]{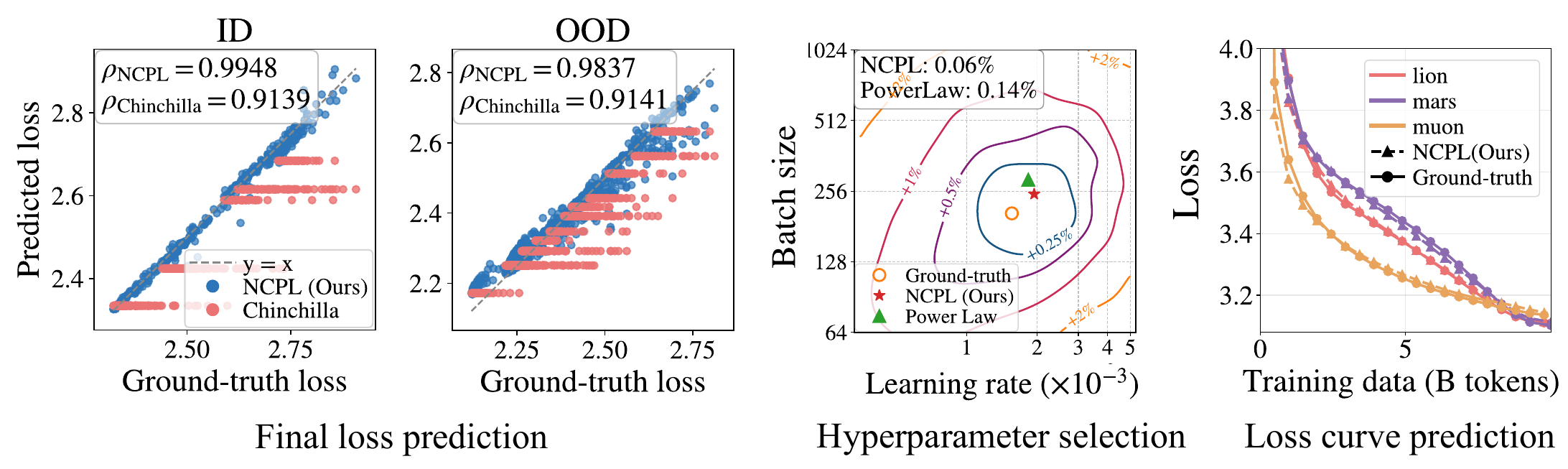}
    \caption{\textbf{An Overview of \predictors\  Performance  Across Tasks. } 
    We split the collected pretraining logs by the model size. In-distribution (ID) means the model size is within the range of the model size in the training set used for \predictor\ and out-of-distribution (OOD) means the model size is larger.
    \textbf{Left: NCPL predicts final loss more accurately than the Chinchilla Law.} Predicted vs. ground-truth loss on validation sets from the StepLaw dataset. The Chinchilla law yields configuration-agnostic prediction whereas \predictor\  takes in the full configuration as inputs and therefore achieves  better prediction. \textbf{Middle: NCPL enables hyperparameter tuning.} Optimal learning rate and batch size prediction in an OOD setup (StepLaw dataset; $N=536$M, $D=28.4$B). Configuration-dependent predictions naturally enable joint tuning over these two hyperparameters, achieving comparable performance to hand-designed functional form~\citep{li2025steplaw}.  \textbf{Right: NCPL can predict the entire loss curve beyond a single loss value.} Predicted vs. ground-truth pretraining loss curves under different optimizers on the validation sets (Marin dataset; $N=520$M, $D=10$B). \predictor\  predicts optimizer-specific curve shapes accurately.
    }
    \ificml
    \vspace{-0.18in}
    \else 
    \fi
    \label{fig:cover}
\end{figure*}

This paper proposes to build a more comprehensive scaling law that accurately maps the \textbf{full} training configuration \conf to performance metrics \metrnospace, such as the final pretraining loss, which we call the \emph{Configuration-to-Performance Scaling Law} (CPL). It addresses limitations of the standard scaling law and provides better predictability. Standard scaling implicitly assumes that all the hyperparameters related to the training algorithms are optimally tuned for the existing runs as well as the hypothetical large-scale runs \citep{hoffmann2022chinchilla,porian2024porianlaw}. However, researchers don't always have resources to tune all hyper parameters, at least not optimally. Thus the scaling law implicitly depends on and varies over the hyperparameter scaling strategy~\citep{everett2024scaling,mup,blakeu2025umup,li2025steplaw}. Moreover, some hyperparameters cannot be arbitrarily tuned due to hardware considerations. For instance, batch size needs to be large enough to fully utilize a large compute cluster~\citep{shuai2024scalinglawlanguagemodels,zhang2025cbsscaling,mccandlish2018empirical}.
In contrast, researchers can use a CPL, which maps out how the performance metric depends on the full set of training hyperparameters, to predict optimal hyperparameters at scale under any external constraints. This simply involves maximizing the predicted performance metric over the choice of the hyperparameters, and thus is simpler than building scaling laws for each individual hyperparameter.

A priori, building a CPL seems to be an overly ambitious goal. The relationship between the configuration and performance is difficult, if not impossible, to have a pre-specified functional form like power law. Thus, we propose to use a \textit{neural ansatz}: a neural network that maps \conf to \metrnospace, with parameters learned from data collected across many existing experiments.

It may appear that we won't have sufficient data from expensive training runs to build such a CPL. Fortunately, open-source pretraining studies, such as Marin~\citep{marin2025announcement}, Step Law~\citep{li2025steplaw} and OLMo~\citep{olmo20242}, have recently released much more diverse public pretraining data. Unlike standard scaling law, CPL can benefit from training on the suboptimal runs (in fact, suboptimal runs are required). Moreover, modern foundation models, as the base models for the training of CPL, may encode prior or theoretical understanding of training dynamics, enhancing transferability and reducing the need for massive run logs.

Encouragingly, we find that training CPL is already feasible with current open pretraining logs and foundation models. We finetune Qwen3-1.7B \citep{yang2025qwen3} to predict performance metrics, including the final pretraining loss and the full loss curve, from a training configuration using a regression objective. We train on over 3,000 pretraining logs from two open-source projects, Marin \citep{marin2025announcement} and StepLaw \citep{li2025steplaw}, and obtain a predictor that we call \textit{Neural Configuration-to-Performance Law (NCPL)}. To test the generalization of this method, we split the data into in-distribution (ID) and out-of-distribution (OOD) sets based on model size, and train only on runs with model size below 430M parameters.
NCPL generalizes to OOD runs that use up to 10$\times$ more compute than any run in the training set  (\cref{sec:method}).
It improves over classical methods along multiple axes: 
\begin{enumerate}[itemsep=0pt,leftmargin=*,topsep=1pt]
    \item NCPL learns how configurations affect the final loss, achieving higher accuracy than the Chinchilla scaling law, which only takes $N$ and $D$ as inputs (\Cref{fig:cover}, Left). On the StepLaw dataset, NCPL achieves over 40\% lower MAE than Chinchilla, and on the Marin dataset it achieves over 20\% lower MAE.
    \item NCPL supports joint tuning of multiple hyperparameters. When restricted to learning rate and batch size, NCPL matches the predictive performance of StepLaw~\citep{li2025steplaw}, a hyperparameter scaling law specifically designed for tuning these two hyperparameters ~(\Cref{fig:cover}, Middle).
    \item NCPL can also be extended to predict the full loss curve, not just the final loss. Previously, achieving this typically required hand-designing complex functional forms \citep{tissue2024scalinglawlearningrate,luo2025multipower,li2025functionalscalinglawskernel} (\Cref{fig:cover}, Right).
    \item NCPL qualitatively learns nuanced interactions between hyperparameters, including a rarely noticed interaction between the optimizer choice and weight-decay strength.
\end{enumerate}

As more public training runs are released, the community can build a shared NCPL from pooled data, and users can further fine-tune their own NCPL starting from this shared base.   Looking ahead, an NCPL may ingest orders of magnitude more training runs than any individual human researcher, while possessing principled understanding comparable to that of a human researcher through the knowledge in the base model. 
While an NCPL likely cannot extrapolate to completely unknown scenarios such as runs testing a novel model architecture or a new dataset, one can continue training it with training logs data from the new scenarios, and expect some level of transfer based on the prior knowledge encoded in the network. This may be more data-efficient than the existing approach of building a brand-new scaling law with all new experiments.

%% file: sections/preliminaries.tex
\section{Preliminaries}\label{sec:method-preliminaries}
\ificml
\textbf{Classical scaling laws.}
\else 
\paragraph{Classical scaling laws.}
\fi 
\citet{kaplan2020openaiscaling} observed that the pretraining loss of large language models decreases monotonically and in a predictable manner as the number of model parameters $N$ and the number of training tokens $D$ increase. More specifically, they proposed that the pretraining loss approximately follows a power-law scaling with respect to $N$ and $D$.
Subsequently, \citet{hoffmann2022chinchilla} introduced a revised formulation, known as the \emph{Chinchilla law}:
\ificml
\vspace{-2mm}
\else
\fi
\begin{align}\label{eq:chinchilla}
    \chin(N, D)
    = E + \frac{A}{N^\alpha} + \frac{B}{D^\beta}.
\end{align}
\ificml
\vspace{-7mm}
\else 
\fi

A key ingredient for accurate scaling-law fitting is proper hyperparameter tuning for each pair $(N, D)$ used to fit the law \citep{hoffmann2022chinchilla, porian2024porianlaw}. 

However, the Chinchilla law itself does not provide guidance on how to tune hyperparameters during pretraining, nor does it predict the pretraining loss for suboptimal training configurations. In this work, we address this limitation by modeling the pretraining loss as a function of the full training configuration using a neural network.

\ificml
\textbf{Hyperparameter scaling law.}
\else 
\paragraph{Hyperparameter scaling law.}
\fi 

Hyperparameter selection plays a crucial role in LLM pretraining.
Recent work approaches this problem by fitting parametric functions that map training scale (model and data size) to the optimal choice of hyperparameters \citep{kaplan2020openaiscaling, deepseekai2024deepseekscaling, bjorck2025microsoftlaw, porian2024porianlaw, hu2024minicpm, wang2024meituanlaw, li2025steplaw, zhou2026setlr}.
For example, \citet{li2025steplaw} models the optimal learning rate and batch size as power-law functions of model size and data size, where the optimal batch size only depends on the data size:
\ificml
\vspace{-2mm}
\else 
\fi
\begin{align}
\eta(N, D) &= c\, N^{\alpha} D^{\beta},  \quad
B(D) = d\, D^{\gamma}. \label{eq:steplaw}
\end{align}
\ificml
\vspace{-8mm}
\else 
\fi

Such approaches rely on strong inductive assumptions about the functional form of the parametric scaling laws. In contrast, \predictor\ enables hyperparameter selection without specifying an explicit functional form a priori by modeling the pretraining loss from heterogeneous training logs in a data-driven manner (\cref{sec:method-hyper}).

%% file: sections/methodology.tex
\section{Methodology}
\label{sec:method}

\subsection{Formulation}

Large language model training outcomes are influenced by a wide range of factors, including model size $N$, data size $D$, model architecture, data recipe, optimization algorithms, training hyperparameters, etc. We collectively refer to these factors as the training configuration \confnospace. We study how to learn a predictive model that maps configurations \conf to performance metrics \metr, such as the final pretraining loss. We refer to this question as learning a \textit{Configuration–to-Performance Scaling Law (CPL)}.

Classical scaling laws \citep{kaplan2020openaiscaling, hoffmann2022chinchilla} can be viewed as a restricted special case of CPL, where the input is limited to only two factors in \confnospace, the model size $N$ and data size $D$, and the relationship is assumed to be a power law (e.g., in~\cref{eq:chinchilla}), while many other influential factors are left unmodeled. However, pre-specifying a closed-form functional relationship for the entire high-dimensional and heterogeneous configuration space is extremely challenging, if not impossible, due to complex and nonlinear interactions among hyperparameters.
Motivated by this perspective, we parameterize the CPL using a generic neural network (specifically, a language model) and train it on a large collection of open-source pretraining runs~\citep{hall2025introducingmarin, li2025steplaw}. This yields what we refer to as the \textit{Neural Configuration–to-Performance Scaling Law (\predictor)}.

\subsection{Neural configuration-to-performance scaling law}
\label{sec:method-ncpl}

We now proceed to the concrete design of \predictor, in which we fine-tune a pretrained language model as regressor $f_\theta$ to map full training configurations \conf to training outcomes \metr.

\ificml
\textbf{Input features and prediction targets.}
\else 
\paragraph{Input features and prediction targets.}
\fi
We use the training configuration of each pretraining run as the input features, including:
\begin{enumerate}
[leftmargin=*,itemsep=0pt,topsep=1pt]
    \item A source identifier indicating which open-source training project the run comes from, to account for source-specific factors not explicitly represented by other features (e.g., data recipe).
    \item Model architecture: model size $N$ (in our case, the number of non-embedding parameters), the number of layers, the number of heads, and the hidden dimension.
    \item Data scale: the number of training tokens ($D$).
    \item Optimizer and training hyperparameters: optimizer, peak learning rate, learning-rate schedule, final learning rate after decay, weight decay, batch size, warmup ratio, gradient clipping threshold, and optimizer-specific hyperparameters (e.g., $\beta_1$, $\beta_2$, and $\epsilon$ for AdamW). 
\end{enumerate}
An example input instance is provided in \cref{fig:input-example}. In this work, we show that language models used as regressors can leverage these configuration features to make accurate, configuration-aware performance predictions.

\begin{figure}[h]
\centering
\begin{minipage}{0.5\linewidth}
\begin{tcolorbox}[
  colback=orange!6,
  colframe=orange!60!black,
  boxrule=0.6pt,
  arc=2mm,
  left=4mm,right=2mm,top=4mm,bottom=1.5mm
]
{\ttfamily\footnotesize
source: steplaw\\
data size: \textcolor{red}{25.0}\\
model size: \textcolor{red}{268.0}\\
num layers: \textcolor{red}{8}\\
num heads: \textcolor{red}{16}\\
hidden dim: \textcolor{red}{9552}\\
optimizer: adamw \\
learning rate: \textcolor{red}{0.000977}\\
lr schedule: cosine\\
weight decay: \textcolor{red}{0.1}\\
batch size: \textcolor{red}{960}\\
max\_grad\_norm: \textcolor{red}{1.0}\\
min\_lr: \textcolor{red}{1e-05}\\
max step: \textcolor{red}{127155} \\
final loss: \textcolor{blue}{0.0235}\\
}
\end{tcolorbox}
\end{minipage}
\caption{An illustrative training configuration used as input to the model. \textcolor{red}{Numbers} are embedded with a two-layer MLP, while other text uses standard token embeddings. The \textcolor{blue}{0.0235} value denotes the target label that the model needs to predict. Note that this number is the residual loss with respect to a Chinchilla baseline (\Cref{eq:residual}). Full examples and additional details are provided in \cref{app:detail-data}.}
\label{fig:input-example}
\ificml
\vspace{-0.28in}
\else 
\fi 
\end{figure}

In this paper, we consider two prediction targets: (i) the final pretraining loss and (ii) the pretraining loss at a specified intermediate training step. Predicting intermediate losses allows us to reconstruct the loss curve by querying losses at multiple intermediate steps.

\ificml
\textbf{Architecture.}
\else
\paragraph{Architecture.}
\fi
We parameterize the regressor $f_\theta$ with a language model. Compared to training from scratch, our ablation study shows that fine-tuning a pretrained model yields better performance on datasets with diverse configurations. We therefore adopt fine-tuning as our default approach (Qwen3-1.7B as the base model in our experiments~\citep{yang2025qwen3}; see \cref{sec:exp-ablation} and \cref{tab:ablation}).
Given the serialized input sequence $x$, the model embeds textual and numerical fields differently (\cref{fig:input-example}).
Textual fields (including field names and categorical values such as optimizer type and learning-rate schedule) use the backbone language model's standard tokenizer and token embeddings.
Numerical values (e.g., $N$, $D$, learning rate, and weight decay) are mapped to the model embedding space via a two-layer MLP. We obtain the scalar prediction by applying a linear layer to the last-layer hidden state at the last input position. 
Our approach is different from text-to-text regression~\citep{akhauri2025performancepredictionlargesystems}, which parameterizes any numerical values as a sequence of digit tokens.

\ificml
\textbf{Training.}
\else 
\paragraph{Training.}
\fi
Let $\trainset=\{(\mathrm{C}^{(i)}, \mathrm P^{(i)})\}_{i=1}^n$ denote a collection of pretraining runs, where
$\mathrm{C}^{(i)}$ is the training configuration of run $i$,  and
$\mathrm P^{(i)} \in \mathbb{R}$ is the observed final pretraining loss (or an intermediate pretraining loss for the loss-curve prediction task, see \cref{app:detail} for details).

Rather than predicting the observed loss $\mathrm P^{(i)}$ directly, we train the model to predict the residual relative to a Chinchilla-law baseline \citep{hoffmann2022chinchilla}.
This design biases learning toward configuration-specific effects beyond the coarse dependence on model size and data size, and empirically improves extrapolation across scales (see ablations in \cref{app:ablation-designchoice}). 
Concretely, we first fit a Chinchilla-law baseline $\hatchin (N,D)$ on $\trainset$, which depends only on the number of model parameters $N$ and the number of training tokens $D$.
As described in \cref{sec:method-preliminaries}, for each $(N, D)$, we select the run with the lowest final pretraining loss across different configurations, and fit the Chinchilla-law form (\cref{eq:chinchilla}) to this collection of selected runs.\footnote{When training on multiple sources, we fit the baseline separately per source.}
We emphasize that the baseline is fit using only the training set, so no information from validation runs is leaked.
For each run $i$, 
we define the residual regression target as
\begin{align}
y^{(i)} = \mathrm P^{(i)} - \hatchin(N^{(i)},D^{(i)}). \label{eq:residual}
\end{align}
\ificml
\vspace{-8.5mm}
\else
\fi
\ificml
\predictor\  is trained by minimizing the mean squared error (MSE): 
$\mathcal{L}(\theta)
= \sum_{i=1}^n \left(f_\theta(\mathrm C^{(i)}) - y^{(i)}\right)^2 / n$.
At inference time, we recover the loss prediction by adding the baseline back.
\else
\predictor\  is trained by minimizing the mean squared error (MSE): 
\begin{align*}\mathcal{L}(\theta)
= \sum_{i=1}^n \left(f_\theta(\mathrm C^{(i)}) - y^{(i)}\right)^2 / n.
\end{align*}
\fi
At inference time, we recover the loss prediction by adding the baseline back.

To stabilize training, we adopt a two-stage fine-tuning scheme similar to the LP-FT method~\citep{kumar2022finetuningdistortpretrainedfeatures}: in Stage 1 updates only the two-layer MLP encoder for numerical fields and the linear prediction head, and Stage 2 fine-tunes all model parameters.

\ificml
\begin{figure*}[t]
    \centering
    \includegraphics[width=\linewidth]{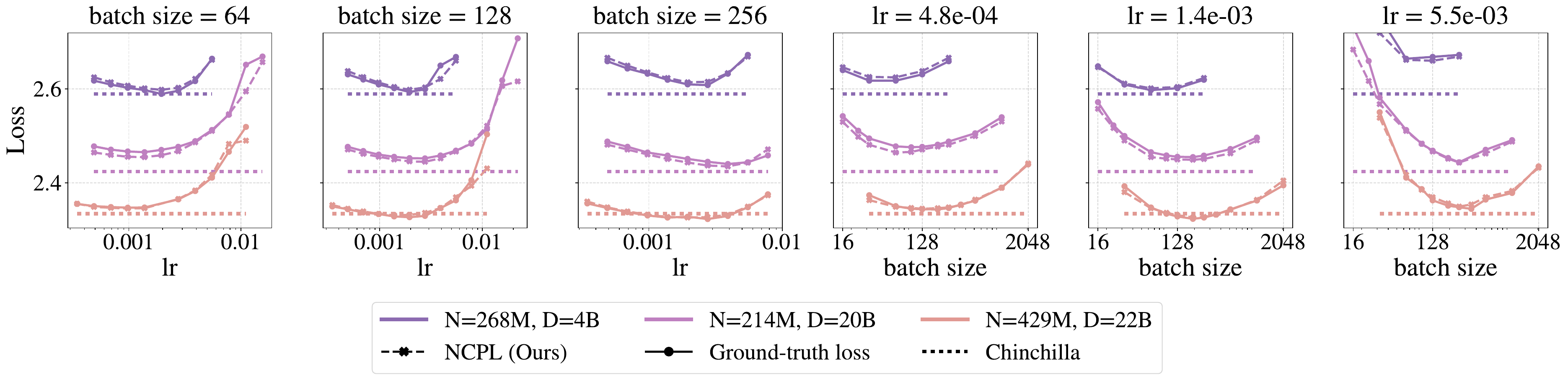}
    \caption{
    \textbf{Predicted vs. ground-truth loss across hyperparameters. 
    }
    Predicted and ground-truth losses are shown across different learning rates and batch sizes for three held-out $(N,D)$ pairs from the StepLaw dataset. Across different $N$ and $D$, \predictor\ accurately predicts how training hyperparameters modulate the final loss, whereas the Chinchilla law yields a single configuration-agnostic prediction for each $(N,D)$ pair. 
    }
    \label{fig:fine-grained-performance}
     \ificml
     \vspace{-0.2in}
     \else 
     \fi
\end{figure*}
\begin{figure}[t]
    \centering
    \includegraphics[width=\linewidth]{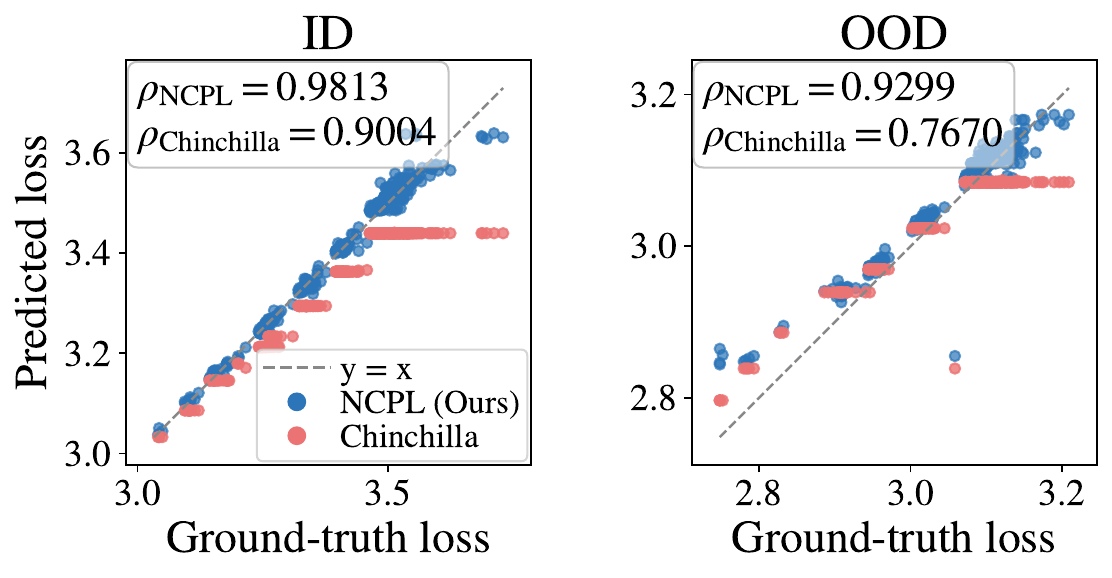}
    \caption{\textbf{Predicted loss vs. ground-truth loss.} 
    Each point of the scatter plots visualizes the predicted vs ground-truth final pretraining loss of an individual training run of a particular training configuration from the Marin dataset (For results of the Steplaw dataset, see \cref{fig:cover} Left).
    \predictor\ takes the full training configurations of individual runs as the input when predicting pretraining loss, whereas the Chinchilla law's prediction only depends on the model size ($N$) and data size ($D$), and therefore predicts the same loss for all runs sharing the same $(N,D)$.
    As a result, \predictor\ achieves substantially higher Spearman correlation $\rho$, as shown in the figure.
    }
    \label{fig:scatter}
    \ificml
    \vspace{-0.3in}
    \else 
    \fi
\end{figure}
\else
\fi

\ificml
\textbf{Evaluation.}
\else 
\paragraph{Evaluation.}
\fi
We report the performance of our \predictor\ and baselines on in-distribution (ID) and out-of-distribution (OOD) validation sets, where the OOD split consists of runs with larger model sizes $N$. The pretraining runs we collected roughly follow the Chinchilla scaling law
.
As a consequence, we are testing the extrapolation of \predictor\ to both larger model size and data size at the same time.

\begin{remark}[Rationale for using language model as the regressors]
Previous work has shown that transformer-based foundation models achieve strong performance as generic regressors (see \cref{sec:related}). Two key advantages are their flexibility in handling heterogeneous inputs (e.g., numerical and categorical features) and their ability to leverage large-scale pretraining on diverse data sources. Looking ahead, the same Transformer backbone readily supports richer prediction targets (e.g., downstream-task performance prediction) and continual fine-tuning. We provide ablation studies against strong tree-based baselines such as XGBoost in \cref{sec:exp-ablation}.
\end{remark}

\subsection{\predictor\ for hyperparameter selection}
\label{sec:method-hyper}

\Predictor\ predicts a loss value from the full training configuration. 
This enables 
selecting multiple hyperparameters jointly
without running expensive training sweeps.
Specifically, for a target model size $N$ and data size $D$, one can enumerate a discrete grid of candidate training configurations and select the configuration that minimizes {\predictor}'s estimated final loss, which serves as a proxy for the true final pretraining loss. 
Experimental results are provided in \cref{sec:exp-main-results}.

\ificml 
\else 
\begin{figure}[t!]
    \centering
    \includegraphics[width=0.6\linewidth]{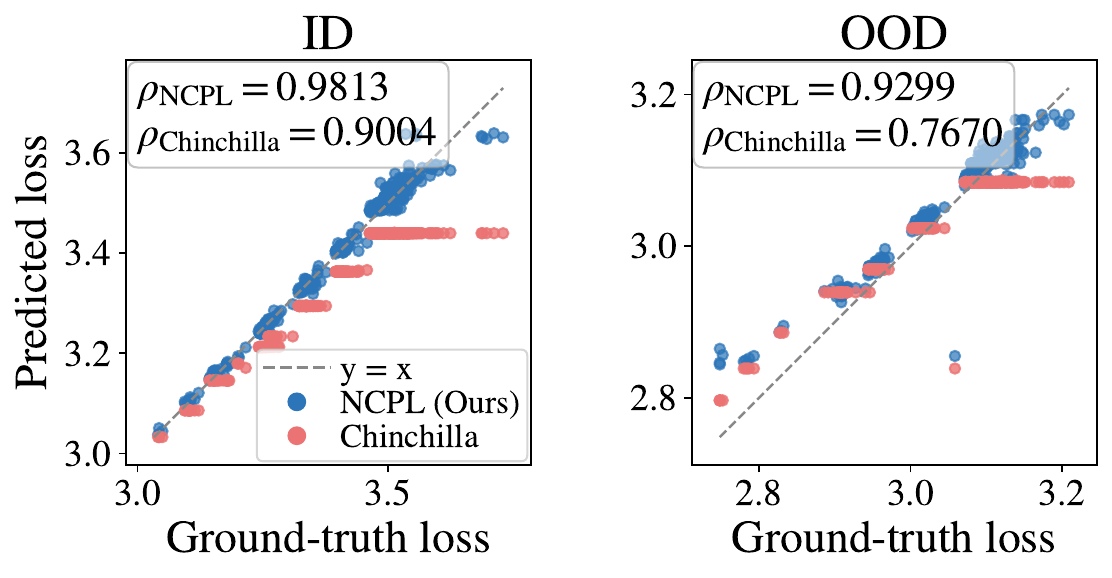}
    \caption{
    \textbf{Predicted loss vs. ground-truth loss.}
        Each point visualizes the predicted vs.\ ground-truth final pretraining loss of an individual run from the Marin dataset (for StepLaw dataset, see \cref{fig:cover} left). \predictor\ uses the full training configuration as input, whereas the Chinchilla law only depends on $(N,D)$ and therefore gives the same prediction for all runs sharing the same $(N,D)$. As a result, \predictor\ achieves substantially higher Spearman correlation $\rho$.
    }
    \label{fig:scatter}
\end{figure}
\fi

%% file: sections/experiments.tex
\section{Experiments}
\label{sec:experiments}

\subsection{Experimental setup}
\label{sec:exp-setup}

\ificml 
\else
\begin{figure}[t!]
    \centering
    \includegraphics[width=0.75\linewidth]{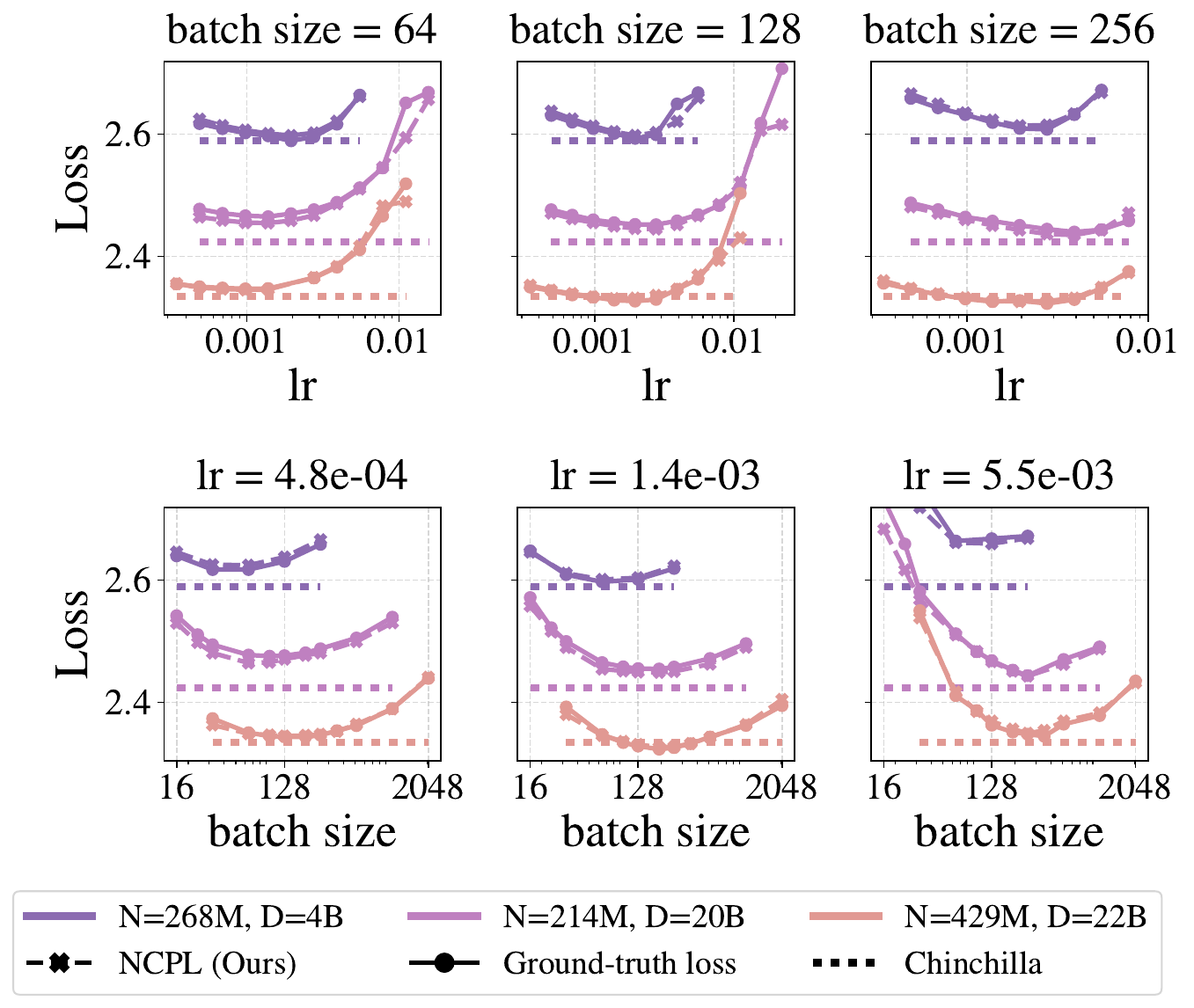}
    \caption{
        \textbf{Predicted vs. ground-truth loss across hyperparameters.}
        Predicted and ground-truth losses are shown across different learning rates and batch sizes for three held-out $(N,D)$ pairs from the StepLaw dataset. Across different $N$ and $D$, \predictor\ accurately predicts how training hyperparameters modulate the final loss, whereas the Chinchilla law yields a single configuration-agnostic prediction for each $(N,D)$ pair.
        }
    \label{fig:fine-grained-performance}
\end{figure}
\fi

\ificml
\begin{figure*}[t]
    \centering

    \begin{subfigure}{0.49\linewidth}
        \centering
        \includegraphics[width=\linewidth]{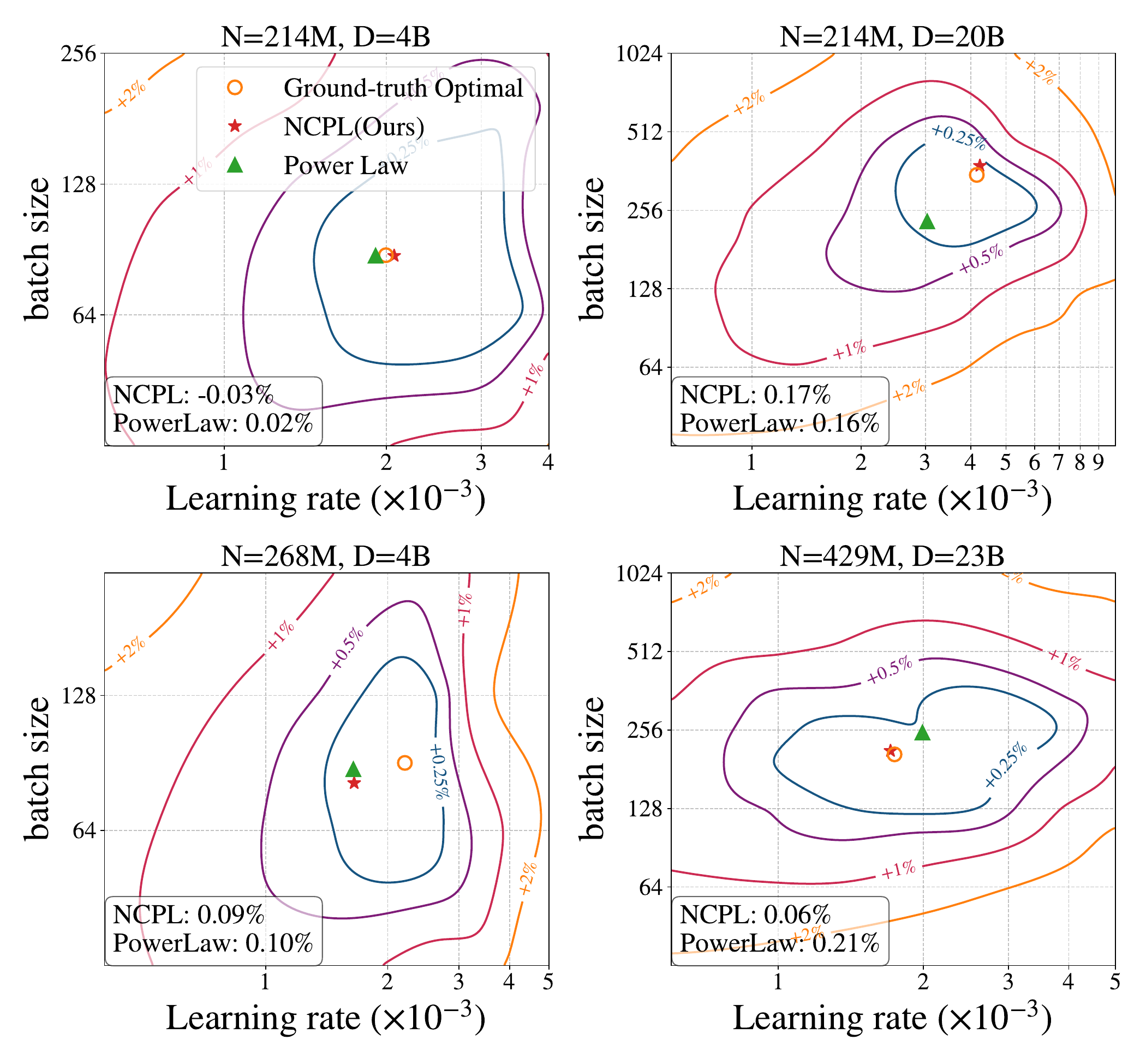}
        \caption{In-distribution}
        \label{fig:contour-id}
    \end{subfigure}
    \hfill
    \begin{subfigure}{0.49\linewidth}
        \centering
        \includegraphics[width=\linewidth]{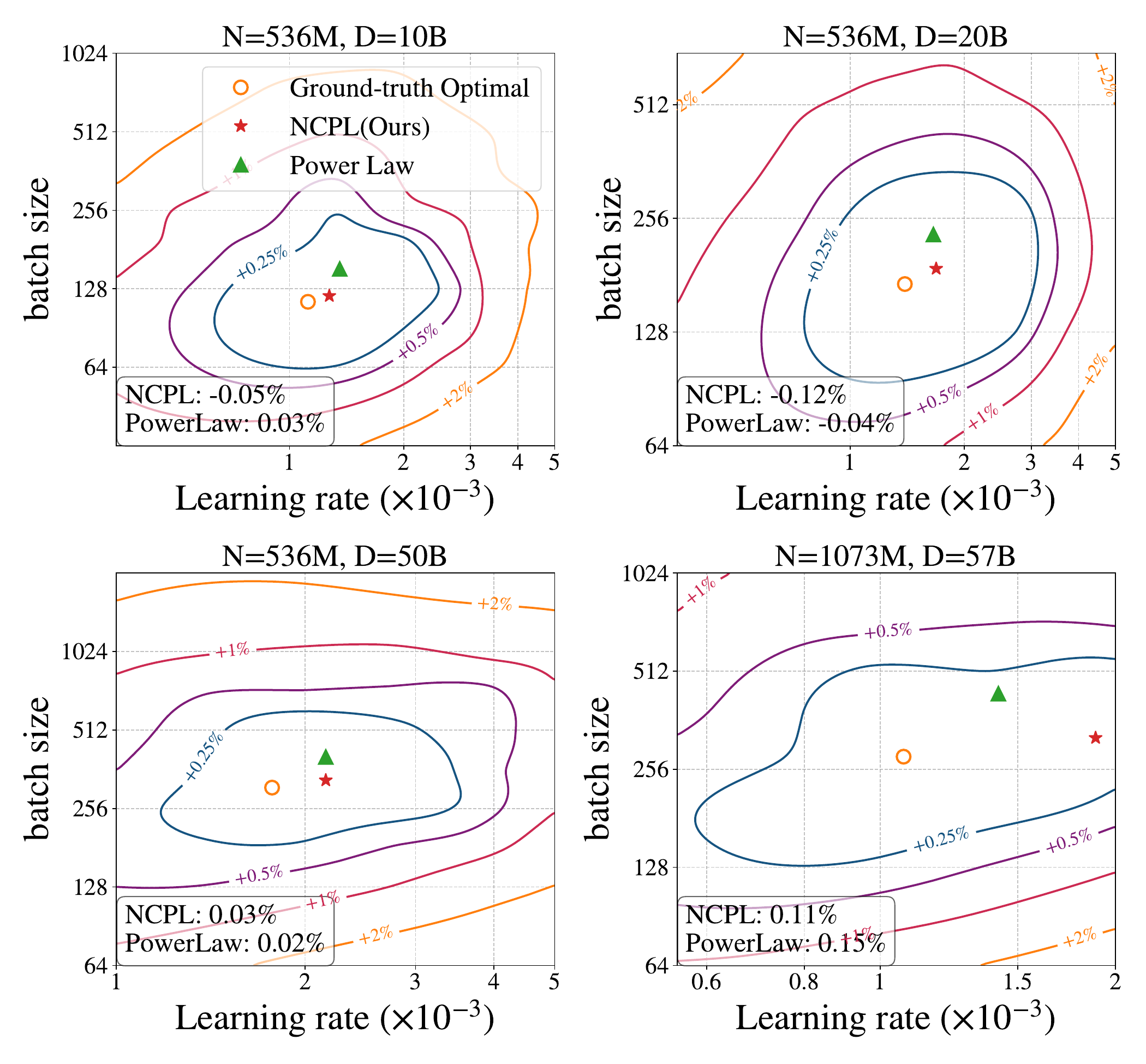}
        \caption{Out-of-distribution}
        \label{fig:contour-ood}
    \end{subfigure}

    \caption{
    \textbf{Hyperparameter selection with \predictor.}
    Predicted optimal learning rates and batch sizes from \predictor\ and the power-law fitting baseline (\cref{eq:steplaw}) on held-out ID and OOD $(N,D)$ pairs.
    The contour lines show the true loss landscape, with labels indicating relative losses to the minimum.
    The bottom legend reports the relative losses of the hyperparameters selected by NCPL and by the power-law baseline.
    \predictors\ prediction aligns with the true optima, and achieves comparable losses to the power-law baseline.
    }
    \label{fig:contour}
    \ificml
    \vspace{-0.26in}
    \else
    \fi
\end{figure*}
\else
\fi

\textbf{Dataset.}
We train and evaluate NCPL on pretraining logs 
collected from two open-source pretraining projects, from which we construct the \emph{Marin Dataset} and the \emph{StepLaw Dataset}, respectively.

\begin{enumerate}[leftmargin=*,itemsep=0pt,topsep=1pt]
    \item \textbf{Marin Dataset.}
    Collected from the Marin Fantastic Optimizers Project \citep{hall2025introducingmarin,wen2025fantasticoptimizers}, which systematically studies the performance and scalability of different optimizers for language model pretraining. The project conducts extensive hyperparameter sweeps across model sizes ranging from 130M to 1.2B parameters and data scales up to 193B tokens.\footnote{Pretraining logs are available at \href{https://wandb.ai/marin-community/optimizer-scaling}{https://wandb.ai/marin-community/optimizer-scaling}.}

    \item \textbf{StepLaw Dataset.}
Collected from the StepLaw Project \citep{li2025steplaw}. The project performs fine-grained sweeps over learning rate and batch size for models ranging from 215M to 1B parameters and data scales up to 56B tokens, while fixing all other hyperparameters and using the AdamW optimizer \citep{loshchilov2019adamw}.The extracted training logs contain model size (with architectural specifications such as number of layers, attention heads, and hidden dimension), data size, learning rate, batch size, pretraining loss. \footnote{Pretraining logs are available at \href{https://wandb.ai/billzid/predictable-scale}{https://wandb.ai/billzid/predictable-scale}.}  %

\end{enumerate}

After excluding runs that are unstable, non-converged, or accidentally terminated (details in \cref{app:detail-data}), we obtain the Marin dataset and StepLaw dataset of 2,549 and 2,581 training logs respectively.
We designate all runs with model sizes larger than 430M parameters as an out-of-distribution (OOD) validation set. 
The remaining runs are split into training and in-distribution (ID) validation sets with an 8:2 ratio.
To make the splits between training set and ID validation set more meaningful and challenging, we randomly split on the level of \emph{(optimizer, model size, data size)} tuples.
Concretely, we first group runs that share the same $(\text{optimizer}, N, D)$ tuple, and then randomly assign each group to either the training set or the ID validation set.
This strategy makes sure that every run in the ID validation set (or the OOD validation set) does not have another run with the same  optimizer, model size, data size in the training dataset. 
In total, the dataset comprises 3,225 training runs, 796 ID validation runs, and 1,109 OOD validation runs.
\footnote{The dataset is available at \href{https://huggingface.co/datasets/zhqwqwq/NCPL-Pretraining-Logs}{https://huggingface.co/datasets/zhqwqwq/NCPL-Pretraining-Logs}.}

\textbf{Model and fine-tuning details.}
We use Qwen3-1.7B as the base model \citep{yang2025qwen3} and adopt the 2-stage training described in~\Cref{sec:method-ncpl}.  More details are provided in \cref{app:detail-train}.

For the main part of the section, \predictors\ target output is the final pretraining loss. However, \predictor\ naturally extends to other objectives, and we also present results in modeling the entire training loss curve.

\ificml
\else
\begin{figure*}[t]
    \centering

    \begin{subfigure}{0.49\linewidth}
        \centering
        \includegraphics[width=\linewidth]{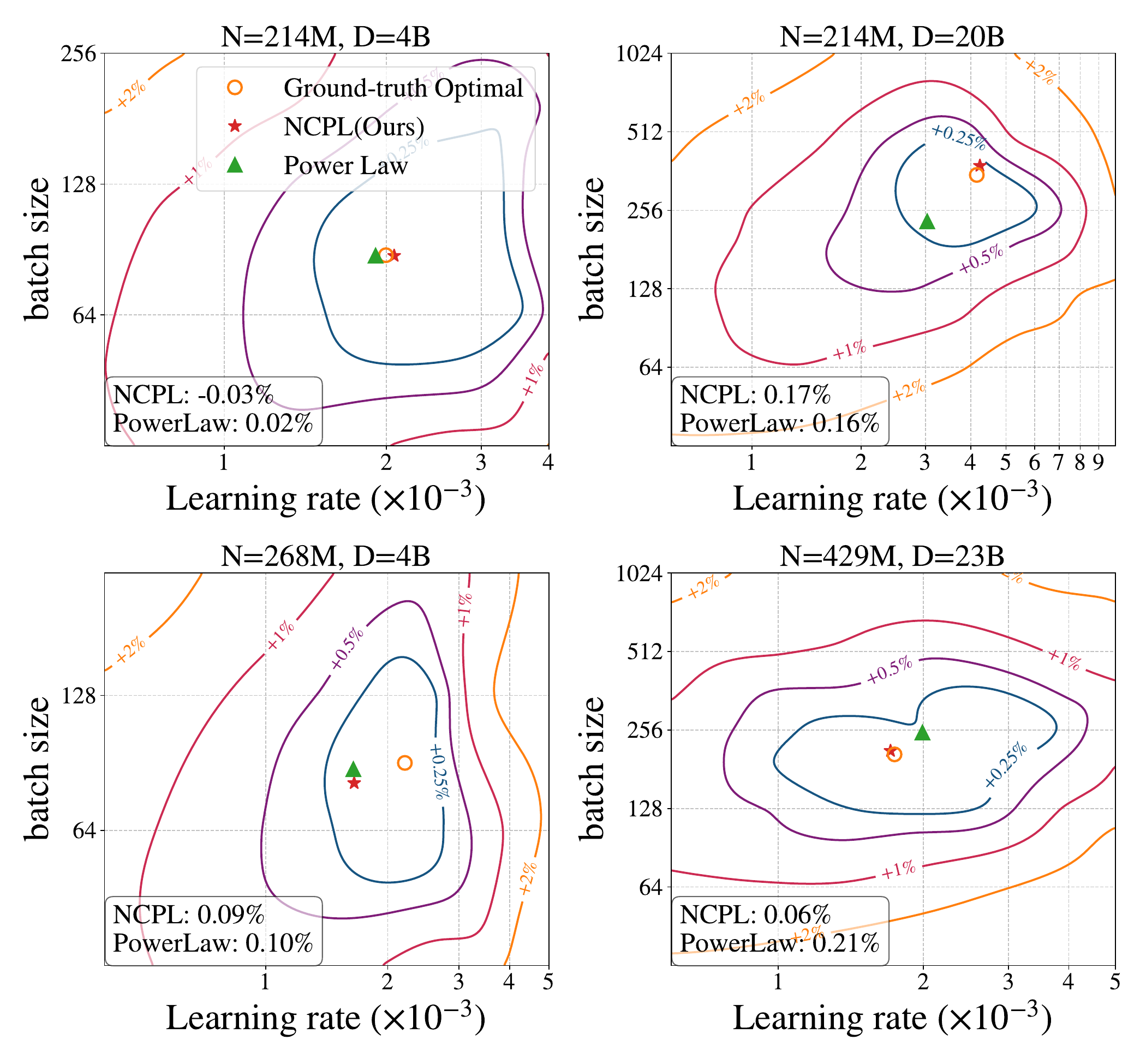}
        \caption{In-distribution}
        \label{fig:contour-id}
    \end{subfigure}
    \hfill
    \begin{subfigure}{0.49\linewidth}
        \centering
        \includegraphics[width=\linewidth]{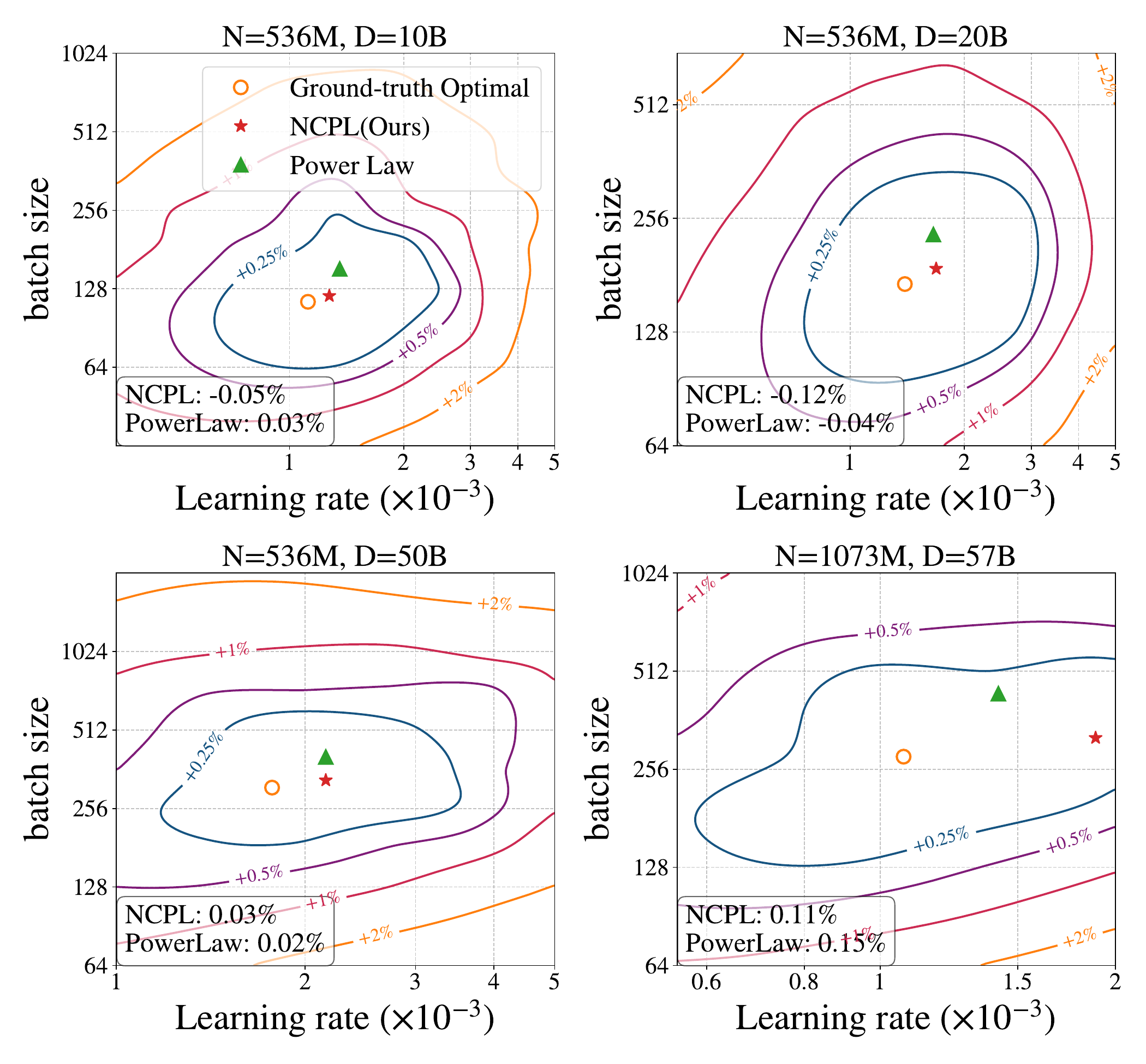}
        \caption{Out-of-distribution}
        \label{fig:contour-ood}
    \end{subfigure}

    \caption{
    \textbf{Hyperparameter selection with \predictor.}
    Predicted optimal learning rates and batch sizes from \predictor\ and the power-law fitting baseline (\cref{eq:steplaw}) on held-out ID and OOD $(N,D)$ pairs.
    The contour lines show the true loss landscape, with labels indicating relative losses to the minimum.
    The bottom legend reports the relative losses of the hyperparameters selected by NCPL and by the power law baseline.
    \predictors\ prediction aligns with the true optima, and achieves comparable losses to the power-law baseline.
    }
    \label{fig:contour}
    \ificml
    \vspace{-0.26in}
    \else
    \fi
\end{figure*}
\fi
\subsection{Main results}\label{sec:exp-main-results}

\ificml
\textbf{Configuration-dependent final loss prediction.}
\else 
\paragraph{Configuration-dependent final loss prediction}
\fi

We first examine \predictors\ ability to predict the final pretraining loss based on the full training configuration. 
In \cref{fig:cover} Left and  \cref{fig:scatter}, each point shows the predicted versus ground-truth final pretraining loss for a single training run of a particular training configuration. 
\predictor\  takes the full training configuration of each run as the input, whereas the Chinchilla Law baseline yields a single configuration-agnostic prediction
for each model-data size pair. Consequently, \predictor\ aligns much more closely with the ground truth, achieving lower prediction errors and higher Spearman correlations than the Chinchilla Law baseline on both ID and OOD validation sets, as summarized in \cref{tab:ablation}. 

Fig~\ref{fig:fine-grained-performance} further visualizes fine-grained predictions over learning rate and batch size sweeps for held-out in-distribution model-data size pairs from the StepLaw dataset. \predictor\ closely tracks the ground-truth U-shaped loss profiles across hyperparameters, illustrating its ability to learn how training hyperparameters affect the final loss.

\ificml
\textbf{Hyperparameter selection.}
\else
\paragraph{Hyperparameter selection.}
\fi
Hyperparameter selection plays a crucial role in large-scale language model pretraining and is a main application of modeling Configuration-to-Performance scaling laws. With the learned \predictornospace, one can sweep over candidate training configurations and use \predictors\ estimated loss as a proxy for the true pretraining loss, thereby identifying optimal hyperparameters without performing expensive training sweeps (see \cref{sec:method-hyper}). 

In this section, we evaluate this capability on held-out ID and OOD model-data size pairs from the StepLaw dataset.
We sweep over \predictors\ predictions to select the optimal learning rate and batch size, and compare its predictions with the power-law scaling rule proposed in \citet{li2025steplaw} (\cref{eq:steplaw}).
As shown in Figure~\ref{fig:contour}, the hyperparameters predicted by \predictor\ closely match the true optima obtained from exhaustive sweeps, in both in-distribution and out-of-distribution settings.
The relative losses are comparable to that of the fitted power-law baseline. Additional results are provided in \cref{app:moreresults}.

We observe that, in the out-of-distribution setting, \predictor tends to suggest larger learning rates, especially for model-data pairs that are farther from the training distribution (e.g., Figure~\ref{fig:contour-ood}, $N{=}1073$M, $D{=}57$B).
We hypothesize that this behavior is because NCPL biases towards its training set of small models which use larger learning rates, and discuss it further in Section~\ref{sec:limitations}.

\ificml
\begin{figure}[t]
    \centering
    \includegraphics[width=\linewidth]{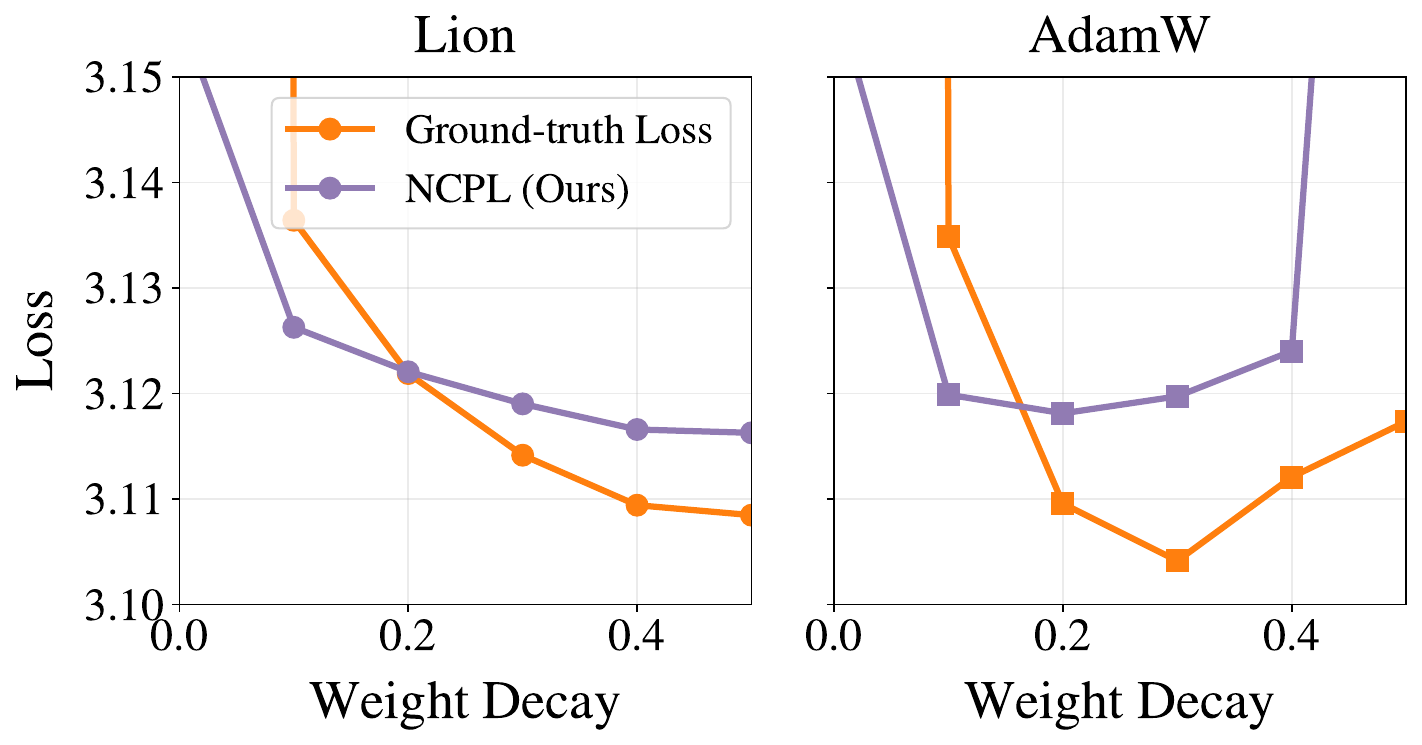}
    \caption{\textbf{\predictor\ learns interactions between weight decay and optimizer choice. 
    }
    It's known that Lion requires substantially larger weight decay than AdamW \citep{chen2023lion, loshchilov2019adamw, wen2025fantasticoptimizers}. 
    \predictor\ predicts this phenomenon on the OOD validation set (Marin Dataset, $N$=520M, $D$=10B).
    }
    \label{fig:lionwd}
    \ificml
    \vspace{-0.38in}
    \else 
    \fi 
\end{figure}
\else 
\fi

\ificml
\else 
\begin{figure}[t]
    \centering
    \begin{minipage}[t]{0.47\linewidth}
        \centering
        \includegraphics[width=\linewidth]{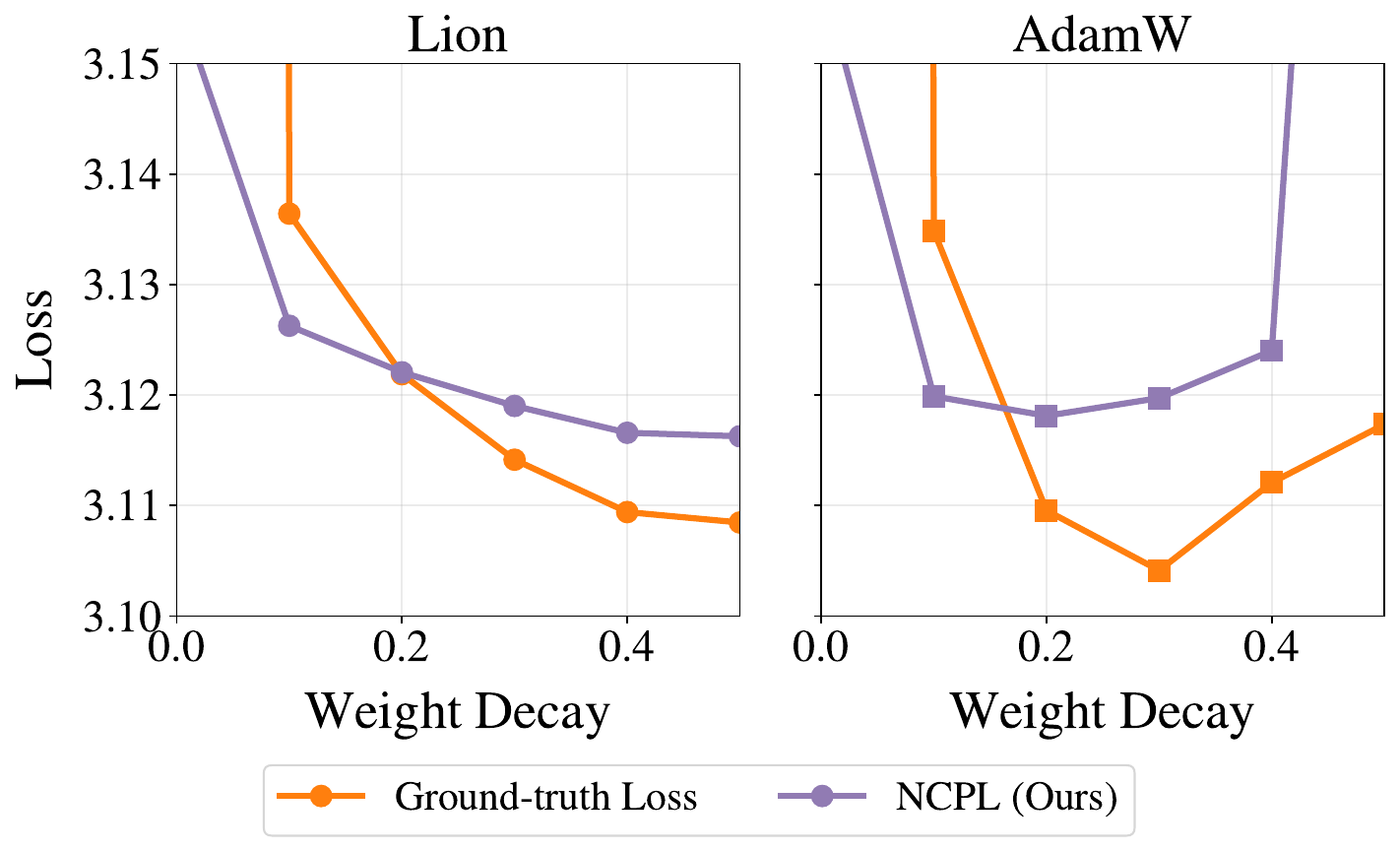}
        \caption{
        \textbf{\predictor\ learns interactions between weight decay and optimizer choice.}
        It's known that Lion requires substantially larger weight decay than AdamW \citep{chen2023lion, loshchilov2019adamw, wen2025fantasticoptimizers}. 
        \predictor\ predicts this phenomenon on the OOD validation sets (Marin Dataset, $N$=520M, $D$=10B).
        }
        \label{fig:lionwd}
    \end{minipage}\hfill
    \begin{minipage}[t]{0.505\linewidth}
        \centering
        \includegraphics[width=\linewidth]{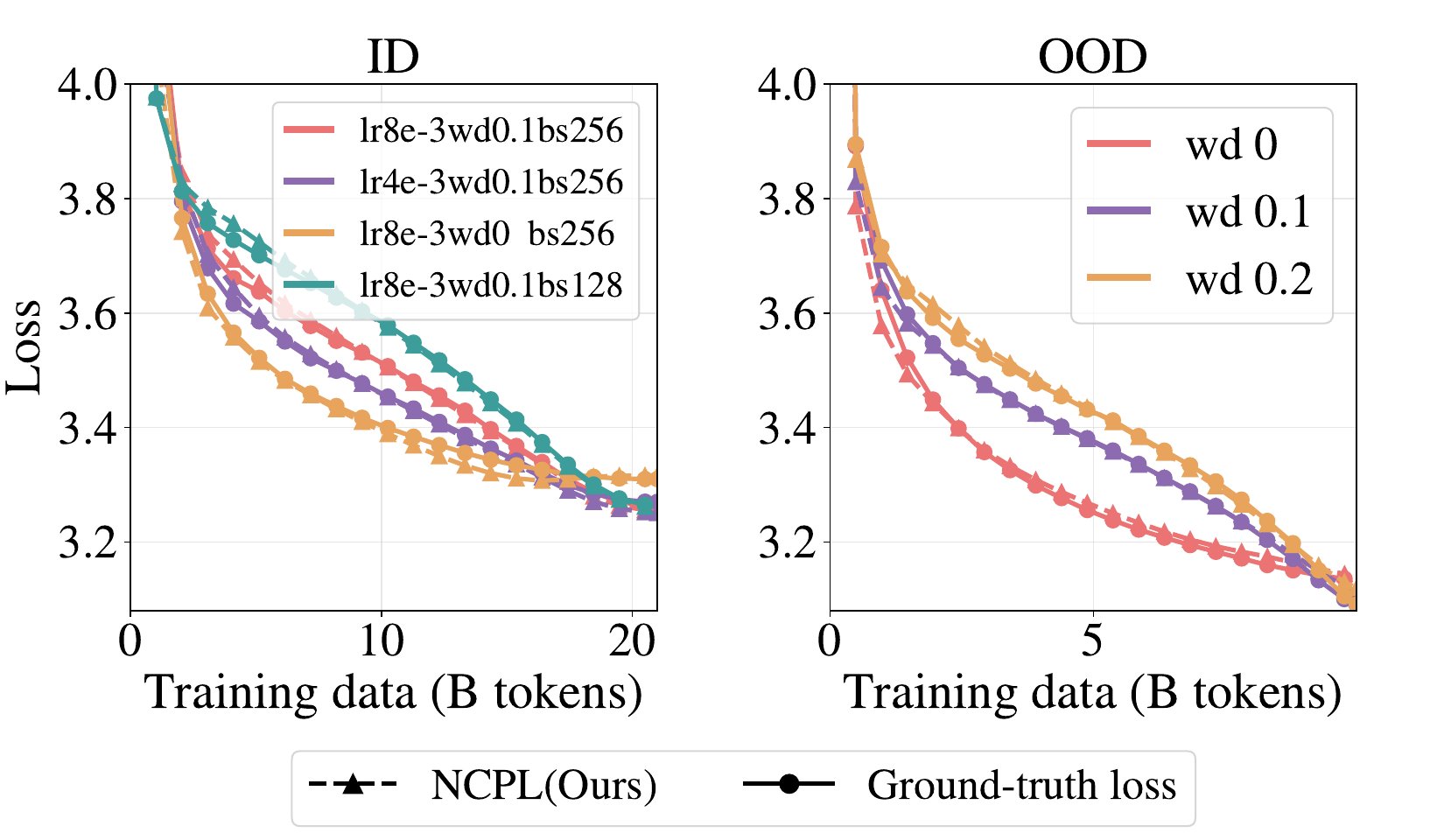}
        \caption{
        \textbf{Loss curve prediction.}
        Ground-truth and predicted pretraining loss curves under different hyperparameter settings on the ID and OOD validation sets. \predictor\ closely tracks the overall trajectories and learns hyperparameter-specific curve shapes. 
        Setting: Marin Dataset.  
        \textbf{Left:} ID validation, $N=130$M, $D=21$B, AdamW optimizer with varying learning rate, weight decay, and batch size.  
        \textbf{Right:} OOD validation, $N=520$M, $D=10$B, Muon optimizer with learning rate $8\times10^{-3}$ and varying weight decay. 
        Results under different optimizers are shown in \cref{fig:cover,fig:losscurve-optimizer}.
        }
        \label{fig:losscurve}
    \end{minipage}
\end{figure}
\fi

\ificml
\textbf{Learning interactions between training configurations.}
\else 
\paragraph{Characterizing interactions between training configurations}
\fi 
Different hyperparameters in the training configuration do not affect final pretraining loss independently; instead, their effects can interact in complicated ways. For example, different optimizers often prefer different hyperparameter choices (e.g., \citet{liu2025moonlight, wen2025fantasticoptimizers, marek2025smallbatch}). By learning a mapping from full training configurations to pretraining performance, we expect \predictor\ to learn such interactions from large-scale pretraining logs.
Here we illustrate this potential with a concrete example. The Lion optimizer \citep{chen2023lion} is known to require substantially larger weight decay than AdamW \citep{loshchilov2019adamw, wen2025fantasticoptimizers}. As shown in \cref{fig:lionwd}, when using larger weight decay (e.g., $\geq 0.4$), AdamW's performance deteriorates markedly, whereas Lion improves. \predictor\ learns this interaction from the training set and generalizes it to the OOD setting.

\begin{table*}[t]
\caption{Comparison of 
various versions of NCPL, XGBoost, and Chinchilla-law baselines for final-loss prediction and loss-curve prediction on both ID and OOD splits. For NCPL, we ablate fine-tuning (\textit{ft}) versus training from scratch (\textit{scratch}) with two backbone sizes (1.7B and 135M). We report mean absolute error (MAE), root mean squared error (RMSE), and Spearman correlation ($\rho$). Overall, NCPL achieves substantially lower error and higher rank correlation than the Chinchilla-law 
by leveraging full training configurations. On StepLaw dataset, where only a small set of hyperparameters vary (learning rate and batch size), \predictor\ (training from scratch) are competitive with \predictor\  with fine-tuning; in contrast, on Marin dataset, which has more diverse configurations, \predictor\  with fine-tuning provides a clear advantage over other variants and baselines. 
}
\centering
\ificml
\small
\else
\scriptsize
\fi
\setlength{\tabcolsep}{4pt}
\begin{tabular}{cccccccccccc}
\toprule
\multirow{3}{*}{\textbf{Data}}
& \multirow{3}{*}{\textbf{Method}}
& \multicolumn{6}{c}{\textbf{Final-loss prediction}}
& \multicolumn{4}{c}{\textbf{Loss Curve Prediction}} \\
\cmidrule(lr){3-8}\cmidrule(lr){9-12}
& & \multicolumn{3}{c}{In-distribution}
& \multicolumn{3}{c}{Out-of-distribution}
& \multicolumn{2}{c}{In-distribution}
& \multicolumn{2}{c}{Out-of-distribution} \\
\cmidrule(lr){3-5}\cmidrule(lr){6-8}\cmidrule(lr){9-10}\cmidrule(lr){11-12}
& & MAE$\downarrow$ & RMSE$\downarrow$ & $\rho$ $\uparrow$
& MAE$\downarrow$ & RMSE$\downarrow$ & $\rho$ $\uparrow$
& MAE$\downarrow$ & RMSE$\downarrow$
& MAE$\downarrow$ & RMSE$\downarrow$ \\
\midrule

\multirow{6}{*}{Marin}
& \multicolumn{1}{c}{NCPL (ft, 1.7B)}
& \textbf{0.0109} & \textbf{0.0169} & \textbf{0.9813}
& \textbf{0.0168} & \textbf{0.0239} & 0.9299
& \textbf{0.0252} & \textbf{0.0475}
& \textbf{0.0363} & \textbf{0.0644} \\
& \multicolumn{1}{c}{NCPL (scratch, 1.7B)}
& 0.0119 & 0.0171 & 0.9774
& 0.0207 & 0.0301 & 0.9324
& 0.0289 & 0.0554
& 0.0386 & 0.0707 \\
& \multicolumn{1}{c}{NCPL (scratch, 135M)}
& 0.0140 & 0.0211 & 0.9777
& 0.0266 & 0.0347 & \textbf{0.9421}
& 0.0306 & 0.0527
& 0.0376 & 0.0713 \\
& \multicolumn{1}{c}{XGBoost}
& 0.0188 & 0.0277 & 0.9754
& 0.0325 & 0.0375 & 0.9227
& 0.0305 & 0.0541
& 0.0385 & 0.0667 \\
& \multicolumn{1}{c}{Chinchilla Law}
& 0.0566 & 0.0676 & 0.9004
& 0.0240 & 0.0326 & 0.7670
& -- & --
& -- & -- \\
& \multicolumn{1}{c}{Per-optimizer Chinchilla}
& 0.0416 & 0.0575 & 0.8268
& 0.0211 & 0.0278 & 0.8062
& -- & --
& -- & -- \\
\midrule

\multirow{5}{*}{StepLaw}
& \multicolumn{1}{c}{NCPL (ft, 1.7B)}
& 0.0097 & 0.0158 & 0.9948
& 0.0223 & 0.0345 & 0.9837
& 0.0278 & 0.0981
& 0.0415 & 0.1696 \\
& \multicolumn{1}{c}{NCPL (scratch, 1.7B)}
& 0.0090 & 0.0150 & \textbf{0.9956}
& 0.0225 & 0.0313 & 0.9876
& \textbf{0.0258} & \textbf{0.0941}
& 0.0384 & 0.1709 \\
& \multicolumn{1}{c}{NCPL (scratch, 135M)}
& \textbf{0.0082} & \textbf{0.0147} & 0.9953
& \textbf{0.0199} & \textbf{0.0284} & \textbf{0.9910}
& 0.0265 & 0.0943
& \textbf{0.0376} & \textbf{0.1666} \\
& \multicolumn{1}{c}{XGBoost}
& 0.0095 & 0.0162 & 0.9947
& 0.0246 & 0.0332 & 0.9863
& 0.0275 & 0.1068
& 0.0471 & 0.1834 \\
& \multicolumn{1}{c}{Chinchilla Law}
& 0.0704 & 0.0949 & 0.9139
& 0.0440 & 0.0670 & 0.9141
& -- & --
& -- & -- \\
\bottomrule
\end{tabular}
\ificml
\vspace{-0.2in}
\else 
\fi
\label{tab:ablation}
\end{table*}

\ificml
\textbf{\predictor\ for loss curve prediction.}
\else 
\paragraph{\predictor\ for loss curve prediction}
\fi 

Beyond accepting the full training  configuration
as inputs, the generality of \predictor\ also allows us to target other metrics beyond final pretraining loss.
We train a variant of NCPL to perform loss curve prediction.
Specifically, we train \predictor\ to predict the pretraining loss at a specified intermediate training step, and reconstruct the loss curve by querying losses at  multiple intermediate steps.
\cref{fig:losscurve} demonstrates the true loss curves and \predictors\ predictions from the Marin Dataset. \predictor\ accurately
predicts the loss curve under different  hyperparameter choices on both ID and OOD generalization settings. This task previously required hand-designing complex multi-component power laws~\citep{luo2025multipower,tissue2024scalinglawlearningrate}, and we show that the shapes of loss curves can be learned directly from data. This also resonates with the recent finding on scaling collapse~\citep{qiu2025superconvergence}, which suggests that the shapes of loss curves are the same across scales up to an affine transformation. 
Moreover, \predictor\  can also predict loss curves for different optimizers; see \cref{fig:cover} and \cref{fig:losscurve-optimizer}.

\ificml
\begin{figure}[!t]
    \centering
    \includegraphics[width=\linewidth]{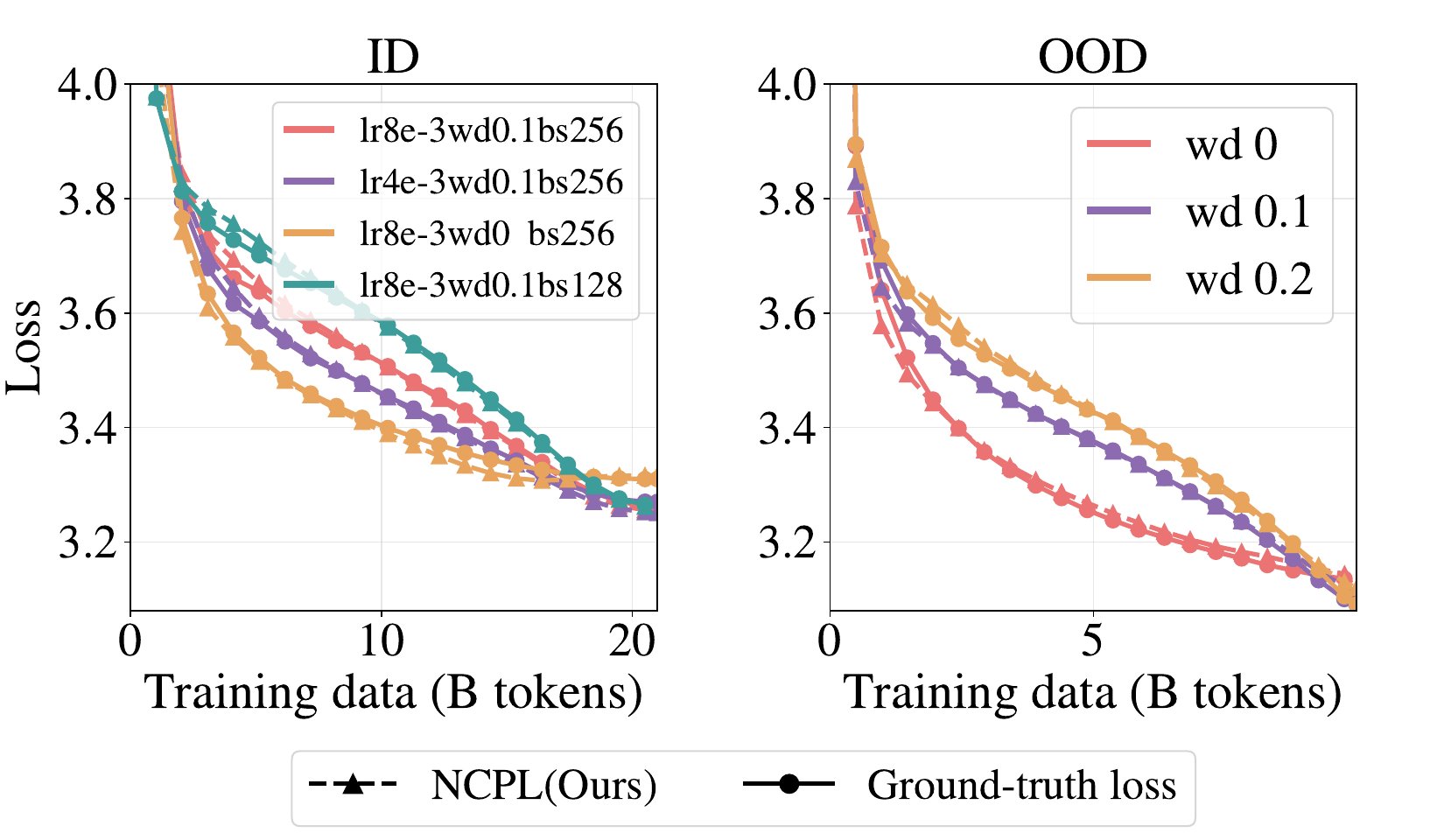}
    \caption{
    \textbf{Loss curve prediction.}
    Ground-truth and predicted pretraining loss curves under different hyperparameter settings on the ID and OOD validation sets. \predictor\ closely tracks the overall trajectories and learns hyperparameter-specific curve shapes.  Setting: Marin Dataset.  \textbf{Left:} ID validation, $N=130$M, $D=21$B, AdamW with varying learning rate, weight decay, and batch size.  \textbf{Right:} OOD validation, $N=520$M, $D=10$B, Muon with learning rate $8\times10^{-3}$ and varying weight decay. Results of loss curve prediction under different opimizers are presented in \cref{fig:cover} and \cref{fig:losscurve-optimizer}.}
    \label{fig:losscurve}
    \ificml
    \vspace{-0.2in}
    \else 
    \fi 
\end{figure}
\else 
\fi

\subsection{Ablation on the Backbone Architecture of \predictor} 
\label{sec:exp-ablation} 

In this section, we ablate alternative instantiations of CPL, including both NCPL and non-neural predictors. Results are reported in Table~\ref{tab:ablation}.
We compare NCPL with XGBoost~\citep{chen2016xgboost} and Chinchilla Law \citep{hoffmann2022chinchilla} baselines on final-loss prediction and loss curve prediction tasks. For NCPL, we further ablate fine-tuning versus training from scratch with different model sizes, using the Qwen3-1.7B~\citep{yang2025qwen3} and SmolLM2-135M~\citep{allal2025smollm2} architectures.

NCPL achieves substantially lower prediction error and higher Spearman correlation than the configuration-agnostic Chinchilla-law baseline by leveraging the full training configuration, rather than only $(N,D)$. 
On the StepLaw dataset, where the configuration variation is restricted to a small set of continuous factors ($N$, $D$, learning rate, and batch size), training from scratch can outperform fine-tuning, and XGBoost achieves comparable performance. In contrast, on the Marin dataset, where many heterogeneous fields vary jointly (e.g., optimizer choices and a wide range of hyperparameters), fine-tuning yields a clear advantage over both training from scratch and XGBoost. Therefore, we adopt the fine-tuned NCPL as our main method.

%% file: sections/related_work.tex
\section{Related work}
\label{sec:related}
\ificml
\textbf{Scaling law.}
\else 
\paragraph{Scaling law.}
\fi

\citet{kaplan2020openaiscaling, hoffmann2022chinchilla} showed the pretraining loss of the Transformer-based LLMs follows a power-law relation with the model size and data size. Subsequent work explores refined fitting protocols and functional forms %
\citep{porian2024porianlaw, li2025predictablescale2, caballero2023brokenneuralscalinglaws}, different model architectures \citep{pmlr-v235-ludziejewski24a, wang2024meituanlaw}, incorporating other training hyperparameters such as learning rate \citep{tissue2024scalinglawlearningrate, xie2024optimization, luo2025multipower,li2025functionalscalinglawskernel} and loss curve prediction \citep{luo2025multipower, tissue2024scalinglawlearningrate, qiu2025superconvergence}. 
\ificml
\else 
Scaling law is also applied 
to broader domains and settings, including  multimodal models \citep{henighan2020scalinglawsautoregressivegenerative} , hyperparameter optimization \citep{kadra2023scaling, pmlr-v139-hashimoto21a}, data mixture \citep{jain2024scaling}, adversarial attacks \citep{liu2025scalinglawsblackbox}, transfer learning\citep{hernandez2021scaling}, etc.
Recent works extends scaling laws to model downstream tasks performance rather than the pretraining loss while some other works point out practical challenges~%
\citep{gadre2025language, ruan2024observational, bhagia2024establishing, khatri2025artscalingreinforcementlearning,lourie2025unreliable}.
\fi
Scaling laws are also 
extended to predict the optimal hyperparameters with the scale of training resources \citep{deepseekai2024deepseekscaling, li2025steplaw, powerline, bjorck2025microsoftlaw, marek2025smallbatch, li2025efficient, zhou2026setlr}.
However, these approaches fit only a small subset of hyperparameters, making it difficult to ensure that the remaining hyperparameters stay optimal during fitting or extrapolation; moreover, they do not model non-parametric configuration choices such as the optimizers.

Recently, \citet{lin2025scalinglawagent} proposes to use advanced LLMs in a scaffold to propose the functional form of scaling laws under different setups. However, identifying the correct functional form mapping from the \textit{entire} configuration \conf to performance \metr can be extremely difficult due to the complicated interactions of different factors. 
We adopt a different methodology to directly use LLM as the regressor to predict the performance from training configurations.

\ificml
\else 
\paragraph{Theory-motivated studies of hyperparameter effects and transfer.}
As complement to data-driven scaling-law fitting, an orthogonal approach studies the effect of hyperparameters and its transfer across training scales through theoretical lens.
Using SDEs as a proxy for stochastic gradient optimization, prior work has proposed scaling rules for how the learning rate should grow with batch size \citep{jastrzkebski2017sgdsde, Adamlrbsscaling}.
The critical batch size, beyond which large-batch training stops yielding proportional efficiency gains, can be characterized via the gradient-noise-scale proxy \citep{mccandlish2018empirical}, and its scaling behavior has been further analyzed under simplified assumptions \citep{zhang2025cbsscaling}.
\citet{wang2025wdscaling} interpret model weights as an exponential moving average of recent updates, and use this view to derive practical scaling rules for weight decay as model and data size grow.
$\mu$P tackles how optimal learning rate can be transferred from smaller to larger models \citep{mup}. Later works extended from the original width scaling setup to depth scaling and other architectural variants \citep{bordelon2024depthwise, dey2025completep, blakeu2025umup, everett2024scaling}. 
However, $\mu P$ primarily focuses on model scaling and does not directly address learning-rate transfer under data scaling, and offers limited guidance for hyperparameters beyond the learning rate.
Recent work argues weight decay plays an important role in stabilizing the update dynamics and thus facilitating learning rate transfer \citep{kosson2025wdimportant}, and propose joint transfer principle of learning rate and weight decay \citep{fan2025properwd}.
While these works offer valuable insights into how hyperparameters influence pretraining performance and provide practical tuning and transfer guidelines, they typically address only a limited subset of hyperparameters.
\fi

\ificml
\textbf{Foundation models as regressors.}
\else 
\paragraph{Foundation models as regressors.}
\fi 
Transformer-based foundation models have demonstrated strong capability in modeling complex input-output relationships beyond natural language processing. Recent work has shown that pretrained language models can be used as general-purpose regressors by direct prompting pretrained models \citep{vacareanuwords} or through supervised learning \citep{garg2022can,hollmann2022tabpfn,huang2020tabtransformer}. 
Compared to classical regression approaches, foundation models offer several key advantages, including leveraging the semantic meaning of features to perform feature selection \citep{jeongllm}, enabling large-scale pretraining on diverse data sources \citep{hollmann2025accurate}, and supporting  online adaptation fine-tuning \citep{songgeneralregressor}. 
\ificml
\else 
At the same time, recent studies have highlighted potential limitations, including sensitivity to data representation that are irrelevant to the underlying learning task \citep{liu2025robustness}.
Foundation models as regressors have been explored in a wide range of domains, for instance, time-series forecasting \citep{gruver2023large,jintime,das2024decoder,goswamimoment} and performance prediction of complex engineered systems \citep{akhauri2025performancepredictionlargesystems}. 
\fi
Our work focuses on predicting the pretraining outcomes of LLMs from full training configurations. To the best of our knowledge, this setting has not been systematically studied in prior work.

\ificml
\else 

\paragraph{Learning-curve extrapolation and hyperparameter optimization.}
Beyond fitting parametric scaling laws, another line of work aims to \emph{predict the training trajectories} (learning curves) from partially-observed training trajectories, enabling early stopping and grey-box resource allocation in hyperparameter optimization.
Early neural and probabilistic approaches include Bayesian neural networks with specialized learning-curve parameterizations \citep{klein2017learning} and probabilistic rollouts for learning-curve extrapolation across hyperparameter settings using models such as Bayesian RNNs \citep{gargiani2019probabilisticrolloutslearningcurve}.
Ranking-based formulations further learn to rank partially observed curves to guide early termination \citep{wistuba2020learningranklearningcurves}.
More recently, transformer-based Prior-Data Fitted Networks (PFNs) enable fast approximate Bayesian learning-curve extrapolation in a single forward pass \citep{adriaensen2023efficientbayesianlearningcurve}, and extend naturally to freeze-thaw Bayesian optimization via in-context surrogates \citep{rakotoarison2024incontextfreezethawbayesianoptimization}.
These PFN surrogates can also be specialized to optimizer hyperparameters, e.g., Adam tuning \citep{athanasiadis2025tuneadamplease}.
In parallel, Deep Power Laws exploit power-law structure for grey-box HPO and budget allocation \citep{kadra2023scaling}.
A very recent work~\citet{hu2026neuralneuralscalinglaws} tries to train a language model to predict downstream performance with the entire accuracy trajectories from a partial run, using token-level validation loss as input. 
\fi

\ificml
We defer more related works on theoretical motivated hyperparameter transfer and learning curve extrapolation to~\Cref{sec:app_related}.
\else 
\fi

%% file: sections/discussion.tex
\section{Conclusion and Discussion}
\label{sec:limitations}

In this work, we propose to learn the mapping from full training configurations to training outcomes using generic neural networks, by training on large-scale and diverse open-source pretraining logs. We instantiate this idea by fine-tuning a pretrained language model (Qwen3-1.7B \citep{yang2025qwen3}), resulting in the \emph{Neural Configuration--Performance Scaling Law} (NCPL). Empirically, NCPL accurately predicts how configuration choices affect final pretraining performance, supports joint hyperparameter tuning and loss-curve prediction, and qualitatively reveals interaction effects among configurations.

Despite these promising results, our current study should be viewed as a proof of concept due to the limited accessibility of open-source pretraining logs. In particular, our training dataset only includes models with up to 430M parameters and OOD validation set only includes models with at most 1.2B parameters. 
Moreover, the diversity of the configurations in the pretraining logs are limited. For example, in our collected open-source pretraining logs, for AdamW, hyperparameter $\beta_1$, $\beta_2$, and $\epsilon$ are rarely tuned, making it difficult to reliably learn their effects and interactions from the logs; also, hyperparameters are swept over only a few discrete values, making predictions on the unseen values less reliable. 
Another example is that the pretraining logs do not have any MoE models or models with linear attention~\citep{dai2024deepseekmoeultimateexpertspecialization,kimiteam2025kimilinearexpressiveefficient,yang2024gated}, and only contain two choices of pre-training datasets~\citep{li2025datacomplmsearchgenerationtraining}

Looking ahead, we expect NCPL-style predictors to improve with the community’s collective efforts to open-source pretraining experiments spanning diverse setups. Continuously incorporating new data should ultimately enable more reliable prediction for large-scale pretraining.

%% file: appendix/details.tex
\section{Experimental details}
\label{app:detail}

\subsection{Data processing}
\label{app:detail-data}

\paragraph{Prediction targets}
Since the per-step pre-training loss exhibits high variance, we use less noisy metrics. For the Marin dataset, we use the language modeling loss on the English split of the C4 dataset \citep{raffel2020exploring}. For the StepLaw dataset, we use the exponentially smoothed training loss with a smoothing coefficient of $0.99$.

\paragraph{Filtering.}
The open-source runs we collect include diverged or failed runs, so we apply a filtering procedure with the following criteria:
(i) unfinished runs;
(ii) diverged runs, defined as those with final pretraining loss $>4$, or with final loss more than $0.3$ above the best loss achieved at the same data size and model size (a large gap in language-model pretraining);
(iii) unstable runs, defined as those with an average loss slope larger than $0.001$ over any 5\% window of training steps.

\paragraph{Training configuration used.}
In \cref{tab:training-config-fields}, we list all training-configuration fields used as inputs to \predictor. Categorical fields are encoded using the backbone language model’s tokenizer and token embeddings. Numerical values are mapped into the embedding space via a two-layer MLP. For $(\beta_1,\beta_2)$, $\epsilon$, and the preconditioner learning rate of the Kron optimizer, since these values lie in a small discrete set in Marin's ablations, we treat them as categorical and encode them with the standard tokenizer. For $\epsilon$, we additionally apply a negative $\log$ transform.

\begin{table}[t]
\caption{Training configuration fields used as inputs to \predictor. For numerical fields, we report the scaling factor applied before feeding them into the numerical encoder.}
\centering
\small
\renewcommand{\arraystretch}{1.15}
\begin{tabular}{p{0.16\linewidth} p{0.45\linewidth} p{0.14\linewidth} p{0.14\linewidth}}
\hline
\textbf{Field group} & \textbf{Included fields} & \textbf{Field type} & \textbf{Scaling Factor} \\
\hline

\multirow{1}{=}{Source tag}
& Source identifier indicating which open-source training project the run comes from
(e.g., Marin or StepLaw) & Categorical & -- \\
\hline

\multirow{4}{=}{Model architecture}
& Model size $N$ (number of non-embedding parameters, in millions) & Numerical & $0.01\times$ \\
& Number of layers & Numerical & $1\times$ \\
& Number of attention heads & Numerical & $1\times$ \\
& Hidden dimension & Numerical & $0.01\times$ \\
\hline

\multirow{3}{=}{Training scale}
& Number of training tokens $D$ (in billions) & Numerical & $1\times$ \\
& Total number of training steps & Numerical & $0.001\times$ \\
& Current ratio of total training steps (for loss curve prediction) & Numerical & $1\times$ \\
\hline

\multirow{13}{=}{Optimizer and hyperparameters}
& Optimizer & Categorical & -- \\
& Peak learning rate & Numerical & $10^{4}\times$ \\
& Learning-rate schedule & Categorical & -- \\
& Final learning rate after decay & Numerical & $10^{4}\times$ \\
& The ratio between final learning rate and peak learning rate & Numerical & $200\times$ \\
& Weight decay & Numerical & $10^{2}\times$ \\
& Batch size & Numerical & $10^{-1}\times$ \\
& Warmup ratio & Numerical & $10^{-2}\times$ \\
& Gradient clipping threshold & Numerical & $1\times$ \\
& Momentum coefficients $(\beta_1,\beta_2)$ for AdamW & Categorical & -- \\
& Numerical-stability constant $\epsilon$ for AdamW & Categorical & -- \\
& Adam learning rate used inside Muon optimizer & Numerical & $10^{4}\times$ \\
& Block size for the Soap optimizer & Numerical & $0.02\times$ \\
& Preconditioner learning rate for Kron optimizer & Categorical & -- \\
\hline

\end{tabular}
\label{tab:training-config-fields}
\end{table}

\paragraph{Scaling.}
Numerical configuration values can vary by orders of magnitude, yet they are all processed by the same numerical encoder. We therefore apply a fixed scaling to each numerical field when constructing inputs, so that different fields fall into a comparable range. The scaling factors are reported in \cref{tab:training-config-fields}.

Example training samples for final-loss prediction and loss curve prediction are shown in \cref{fig:input-examples}.

\begin{figure}[p]
\centering

\begin{minipage}[t]{0.48\linewidth}
\centering
\begin{tcolorbox}[
  colback=orange!6,
  colframe=orange!60!black,
  boxrule=0.6pt,
  arc=2mm,
  left=4mm,right=2mm,top=4mm,bottom=1.5mm
]
{\ttfamily\footnotesize
source: steplaw\\
data size: \textcolor{red}{25.0}\\
model size: \textcolor{red}{268.0}\\
num layers: \textcolor{red}{8}\\
num heads: \textcolor{red}{16}\\
hidden dim: \textcolor{red}{9552}\\
optimizer: adamw \\
learning rate: \textcolor{red}{0.000977}\\
lr schedule: cosine\\
warmup: \textcolor{red}{2000}\\
weight decay: \textcolor{red}{0.1}\\
batch size: \textcolor{red}{960}\\
beta1: 0.9\\
beta2: 0.95\\
epsilon: {8}\\
max\_grad\_norm: \textcolor{red}{1.0}\\
minlr ratio: \textcolor{red}{0.00102}\\
minlr: \textcolor{red}{1e-05}\\
max step: \textcolor{red}{127155} \\
final loss: \textcolor{blue}{0.0235}\\
\\
\\
}
\end{tcolorbox}
\end{minipage}
\hfill
\begin{minipage}[t]{0.48\linewidth}
\centering
\begin{tcolorbox}[
  colback=orange!6,
  colframe=orange!60!black,
  boxrule=0.6pt,
  arc=2mm,
  left=4mm,right=2mm,top=4mm,bottom=1.5mm
]
{\ttfamily\footnotesize
source: marin\\
data size: \textcolor{red}{26.8}\\
model size: \textcolor{red}{134.0}\\
num layers: \textcolor{red}{32}\\
num heads: \textcolor{red}{8}\\
hidden dim: \textcolor{red}{512}\\
optimizer: soap \\
learning rate: \textcolor{red}{0.0016}\\
lr schedule: cosine\\
warmup: \textcolor{red}{1000}\\
weight decay: \textcolor{red}{0.1}\\
batch size: \textcolor{red}{1280}\\
beta1: 0.95\\
beta2: 0.98\\
epsilon: {15}\\
max\_grad\_norm: \textcolor{red}{1.0}\\
minlr ratio: \textcolor{red}{0}\\
minlr: \textcolor{red}{0}\\
block\_size: \textcolor{red}{256}\\
max step: \textcolor{red}{127155} \\
frac: \textcolor{red}{0.3907} \\
final loss: \textcolor{blue}{0.6609}\\
}
\end{tcolorbox}
\end{minipage}

\caption{Example training samples. Left: an input for final loss prediction. Right: an input for loss curve prediction. \textcolor{red}{Numerical values} are embedded with a two-layer MLP, while other text is embedded using standard token embeddings. The \textcolor{blue}{values 0.0235 and 0.6609} denote target labels and are not part of the input.}
\label{fig:input-examples}
\end{figure}

\subsection{Training setup and hyperparameters}
\label{app:detail-train}

\paragraph{Training NCPL for final loss prediction.} As described in \cref{sec:exp-setup}, we adopt a two-stage training pipeline for stability. We use Qwen3-1.7B as the base model \citep{yang2025qwen3}. In the first stage, we freeze the backbone and update only the two-layer MLP encoder for numerical fields and the linear prediction head.  We train for 20 epochs with learning rate $5\times10^{-5}$ and a warmup ratio of 0.1 of total steps. In the second stage, we fine-tune all model parameters for 1000 epochs using learning rate $1\times10^{-5}$ with a 1000-step warmup. We reset the optimizer state between the two stages. In both stages, we use AdamW with linear learning-rate decay, weight decay $0.01$, and batch size $480$. We fine-tune the model using \texttt{float32} precision.

\textbf{Training \predictor\  for loss curve prediction.}
For each run we uniformly sample up to 30 intermediate checkpoints, append a scalar field indicating the fraction of total training steps completed, and train the model to predict the difference between the current loss and the Chinchilla baseline (\cref{fig:input-examples} Right). We use the same two-stage procedure and hyperparameters, but train for 10 epochs in the first stage and 400 epochs in the second stage.

\subsection{Evaluations}

\paragraph{Hyperparameter selection.}
The contour plots are generated by interpolating the scattered ground-truth loss values in log-scaled learning-rate/batch-size space using a smooth RBF interpolant, followed by light Gaussian smoothing; we then draw iso-loss contour lines on the resulting surface. The ``minimum'' markers for both ground truth and NCPL are obtained by fitting a quadratic surface in log space using only near-optimal points whose loss is within $1\%$ of the minimum predicted loss, and taking the minimizer of the fitted quadratic.

%% file: appendix/additionalresults.tex
\ificml
\else 
\vspace{10mm}
\fi
\section{Additional results}

\subsection{More results of fine-tuned \predictor. }
\label{app:moreresults}

\begin{figure}[p]
    \centering
    \includegraphics[width=\linewidth]{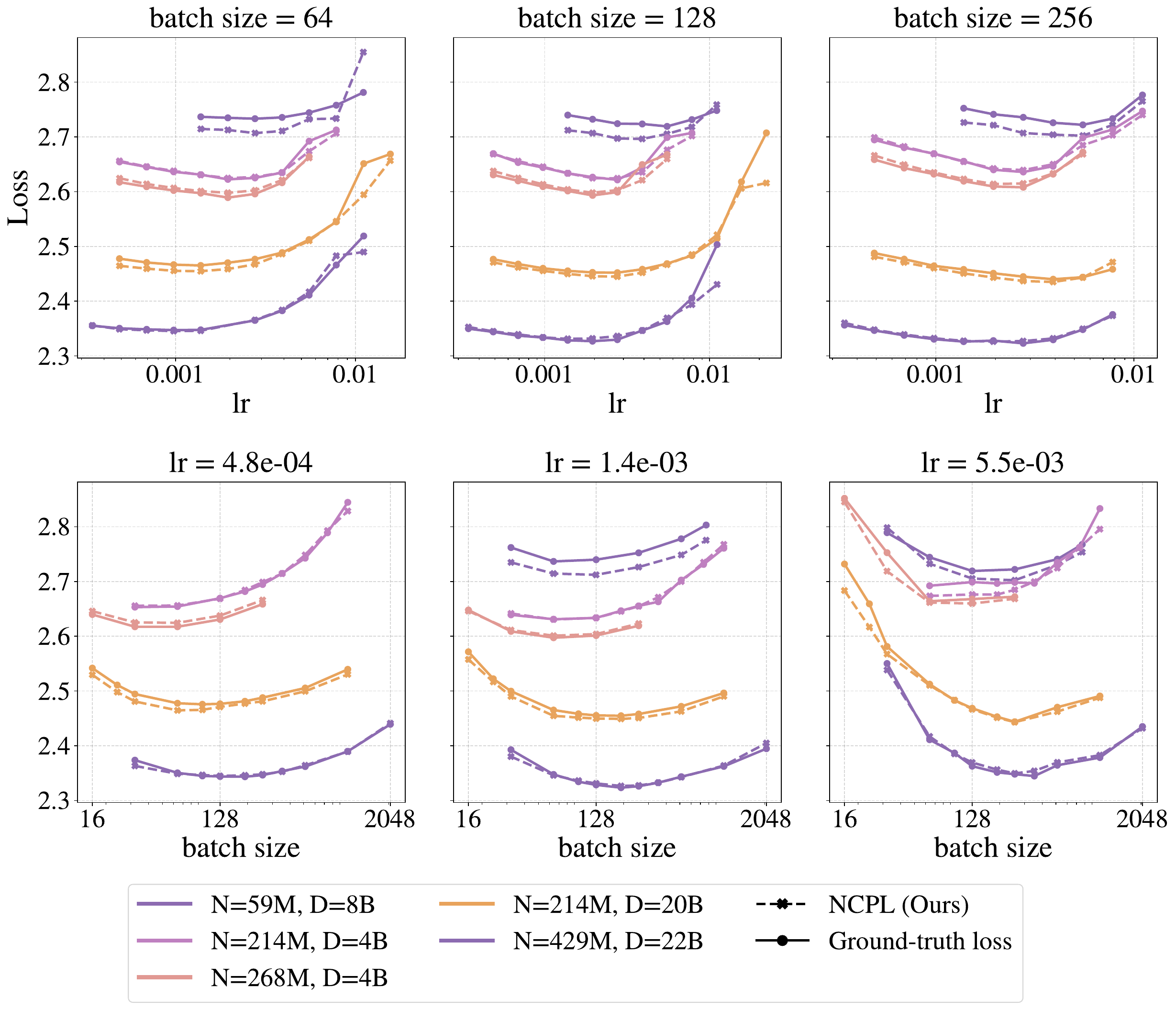}
    \caption{Final-loss prediction across learning rates and batch sizes for all $5$ ID held-out $(N,D)$ pairs. }
    \label{fig:fine-grained-app-id}
\end{figure}

\begin{figure}[p]
    \centering
    \includegraphics[width=\linewidth]{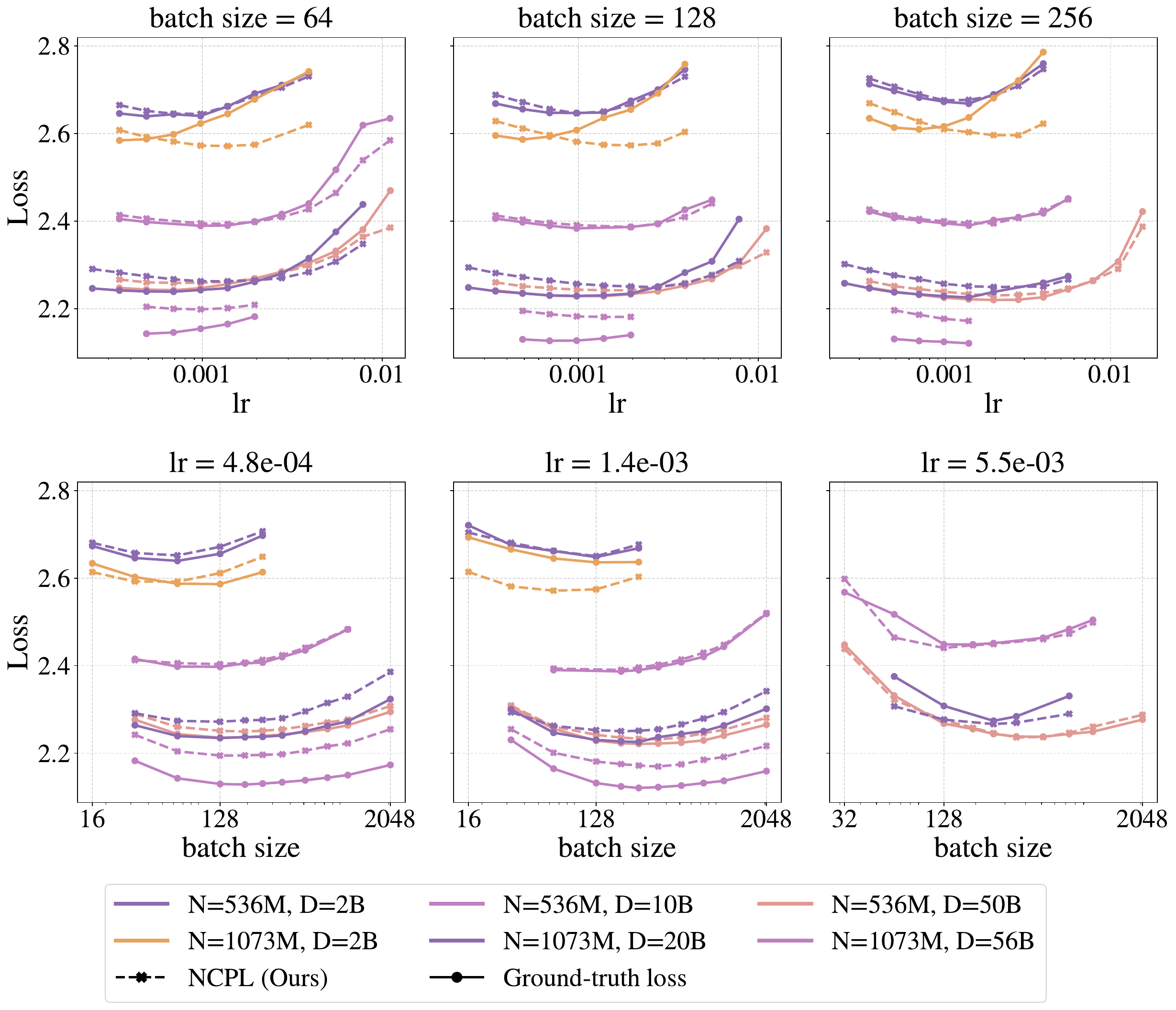}
    \caption{Final-loss prediction across learning rates and batch sizes for $6$ OOD held-out $(N,D)$ pairs. }
    \label{fig:fine-grained-app-ood}
\end{figure}

\begin{figure}
    \centering
    \includegraphics[width=\linewidth]{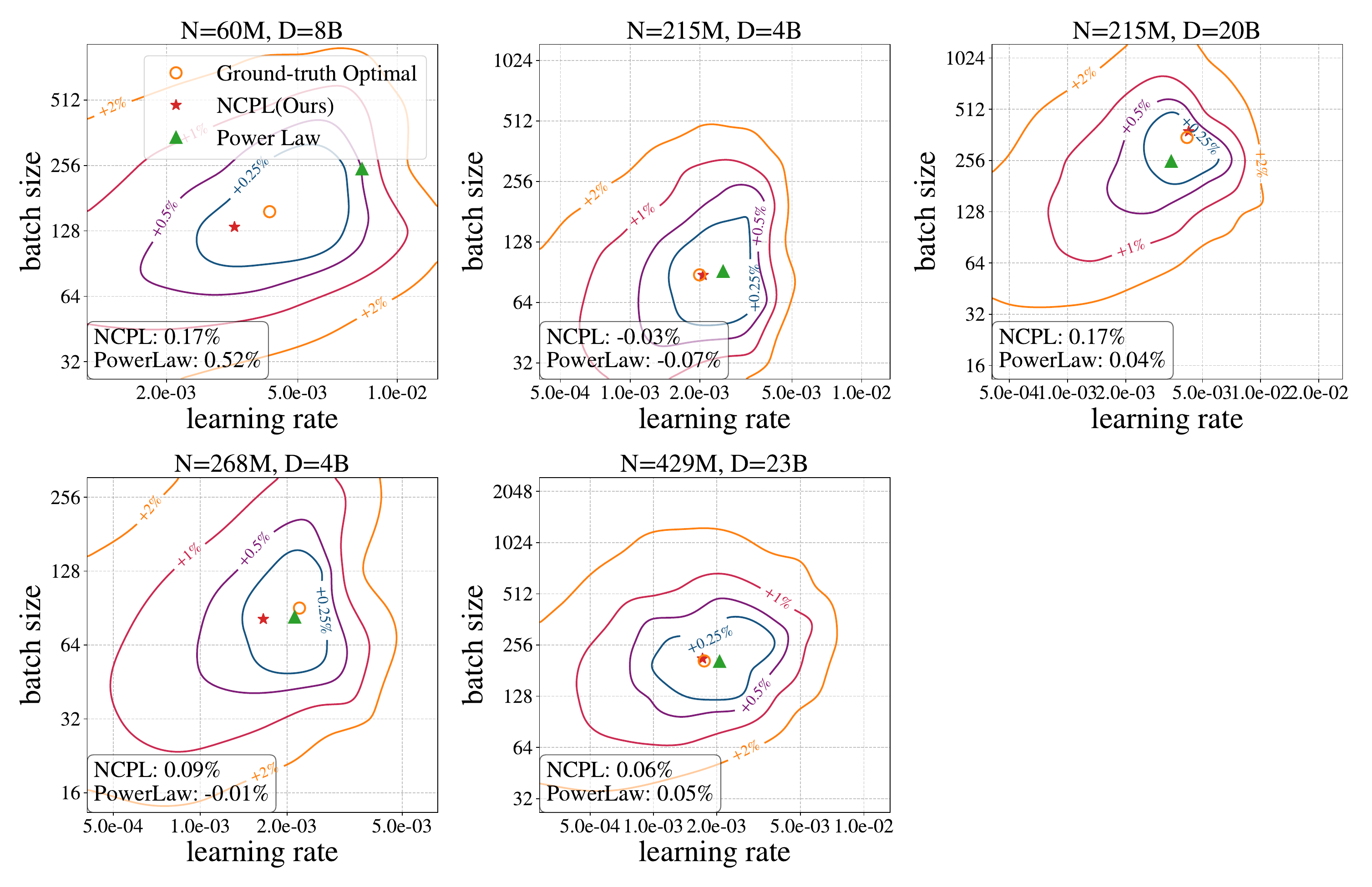}
    \caption{Predicted optimal learning rates and batch sizes for \predictor\  and the power law baseline on all $5$ held-out ID $(N,D)$ pairs, together with their relative losses. }
    \label{fig:contour-all-id}
\end{figure}

\begin{figure}
    \centering
    \includegraphics[width=\linewidth]{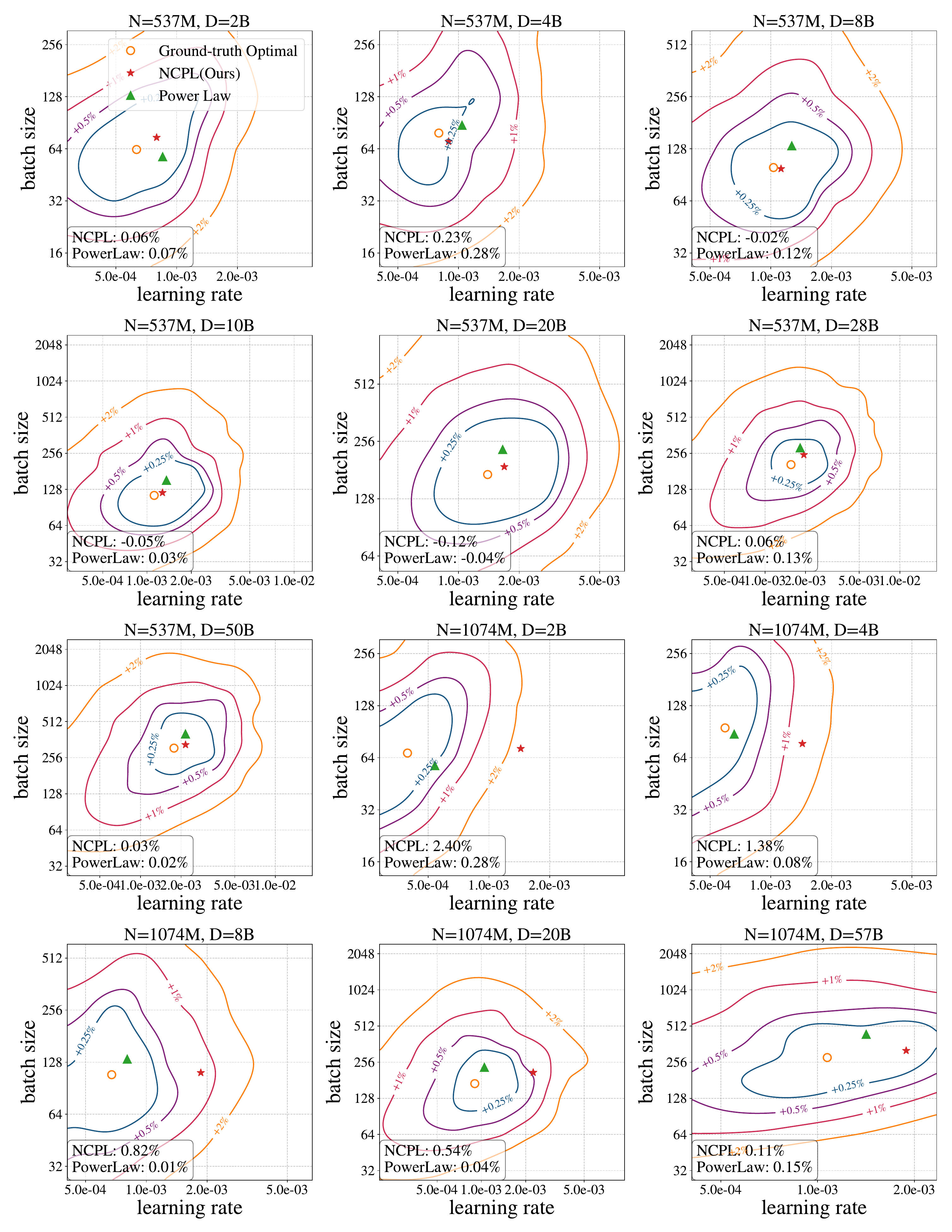}
    \caption{Predicted optimal learning rates and batch sizes for \predictor\  and the power law baseline on all $12$ held-out OOD $(N,D)$ pairs, together with their relative losses. }
    \label{fig:contour-all-ood}
\end{figure}

\begin{figure}
    \centering
    \includegraphics[width=0.95\linewidth]{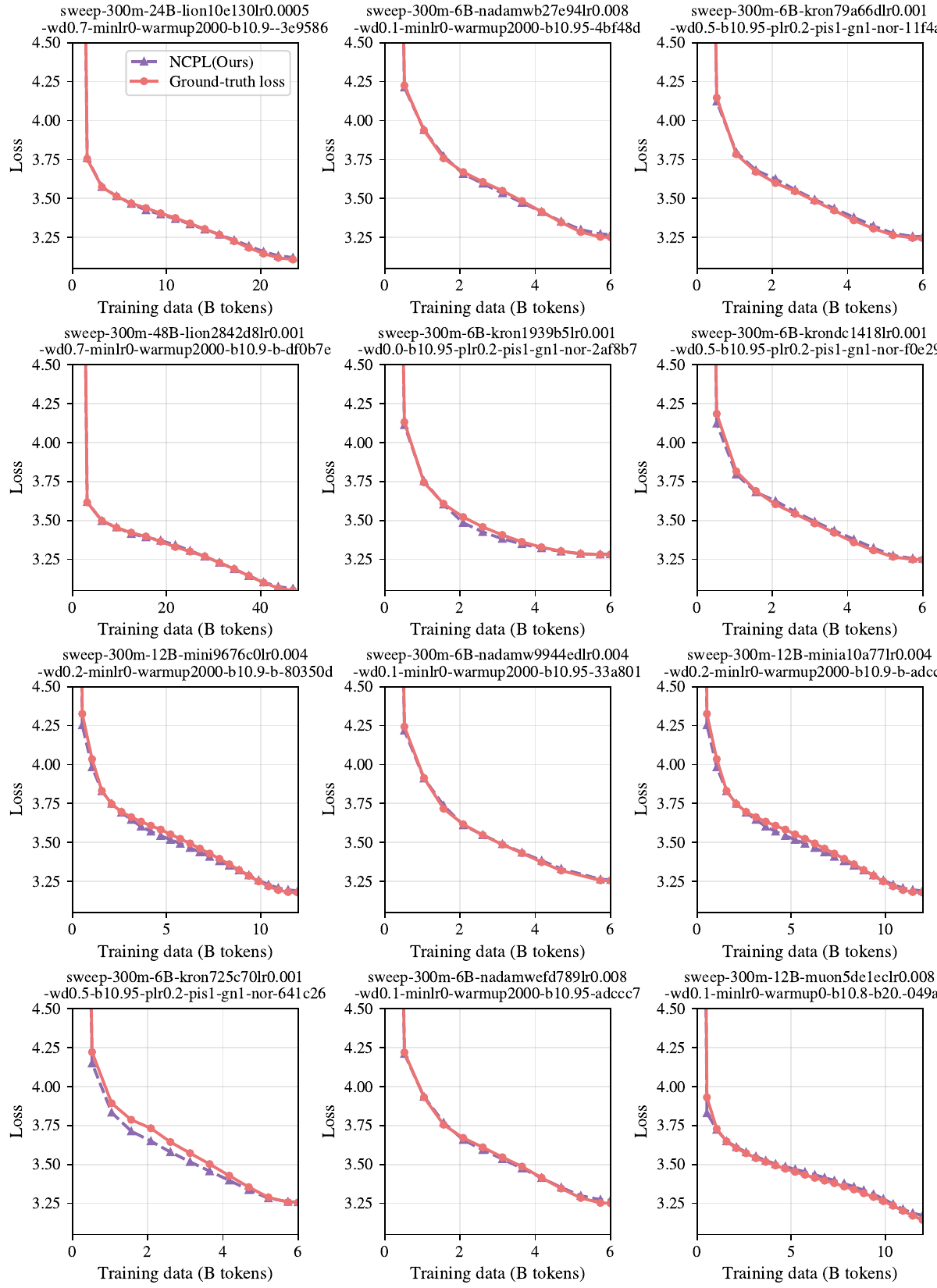}
    \caption{Loss-curve prediction on 12 randomly sampled ID runs (Marin, $300$M). Each subplot title is the corresponding run name from the original Wandb project \citep{wen2025fantasticoptimizers}.}
    \label{fig:curve-12-id}
\end{figure}

\begin{figure}
    \centering
    \includegraphics[width=0.95\linewidth]{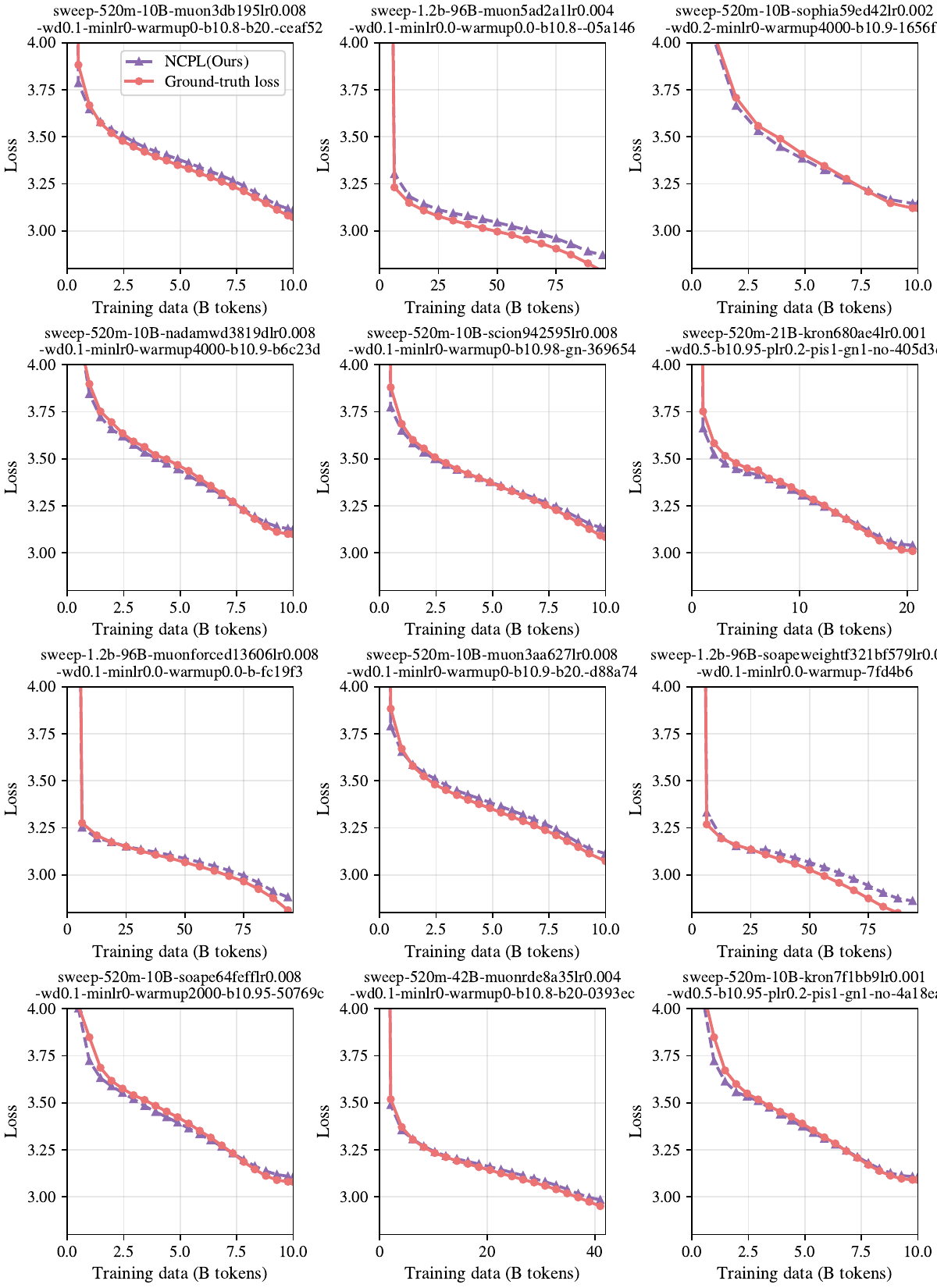}
    \caption{Loss-curve prediction on 12 randomly sampled OOD runs (Marin). Each subplot title is the corresponding run name from the original Wandb project \citep{wen2025fantasticoptimizers}.}
    \label{fig:curve-12-ood}
\end{figure}

\begin{figure}[t]
    \centering
    \includegraphics[width=0.75\linewidth]{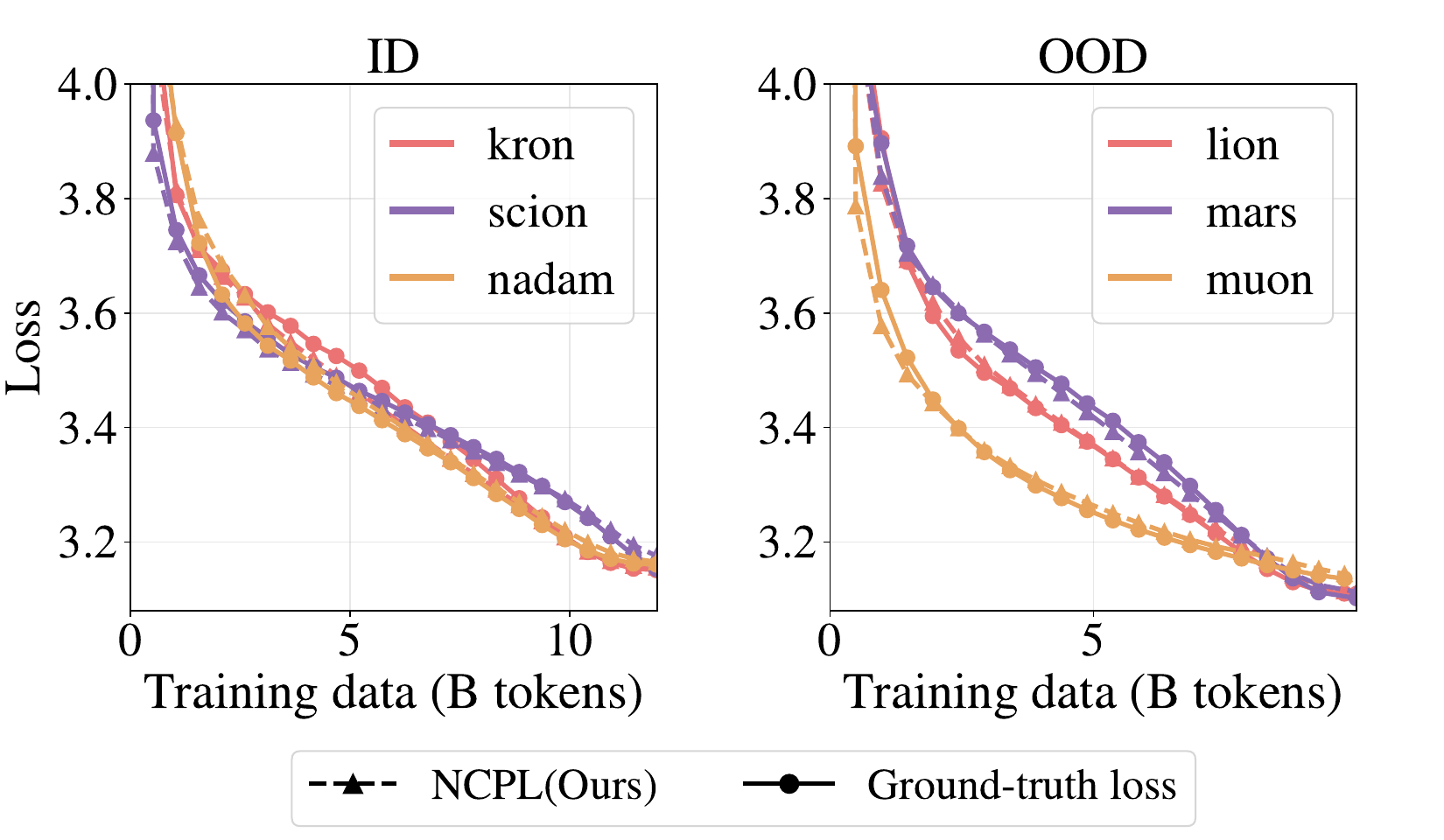}
    \caption{
    \textbf{Loss curve prediction.}
    Ground-truth and predicted pretraining loss curves under different optimizer settings on the ID and OOD validation sets. \predictor\ closely tracks the overall trajectories and captures optimizer-specific curve shapes.  Setting: Marin Dataset.  \textbf{Left:} ID validation, $N=300$M, $D=12$B, optimizers Kron/Scion/Nadam. \textbf{Right:} OOD validation, $N=520$M, $D=10$B, optimizers Lion/Mars/Muon. Results of loss curves under different hyperparameter are shown in \cref{fig:losscurve}.}
    \label{fig:losscurve-optimizer}
\end{figure}

Final pretraining loss prediction results across learning rates and batch sizes on the StepLaw dataset are shown in \cref{fig:fine-grained-app-id} (ID) and \cref{fig:fine-grained-app-ood} (OOD). Hyperparameter selection results for all held-out model–data size pairs are shown in \cref{fig:contour-all-id} and \cref{fig:contour-all-ood}.
At the largest extrapolation scale ($N=1073$M), \predictor's predicted pretraining loss tends to be higher than the ground-truth loss. This is partly because the Chinchilla baseline itself overpredicts at this scale; adding our predicted residual on top of an inflated baseline can further increase the final prediction. In addition, the model predicts a larger optimal learning rate for larger $N$ when the $D/N$ ratio is small. This trend may reflect limited data diversity in the training set, and could be alleviated by incorporating more training logs.
Loss-curve prediction results on the Marin dataset across different optimizers (ID and OOD) are presented in \cref{fig:losscurve-optimizer}. Fig. \ref{fig:curve-12-id} and \cref{fig:curve-12-ood} show randomly sampled loss curve prediction results on ID and OOD runs, without cherry-picking.
\subsection{Ablation on \predictor\ design choices}
\label{app:ablation-designchoice}

We study two key design choices in NCPL: (i) predicting residuals relative to a Chinchilla-law baseline, and (ii) encoding scalar configuration values using numerical tokens via a two-layer MLP. We consider two ablated variants:  predicting the final loss directly, rather than the residual with respect to the Chinchilla baseline, and tokenizing numerical fields with a standard tokenizer, rather than encoding them with a two-layer MLP. Results shown in \cref{tab:ablation-designchoice} show consistent gains from both residual prediction and numerical token encoding.

\subsection{Additional results of  \predictor\  with training from scratch.}
\label{app:additional-tfs}

\begin{figure}[t]
    \centering
    \includegraphics[width=\linewidth]{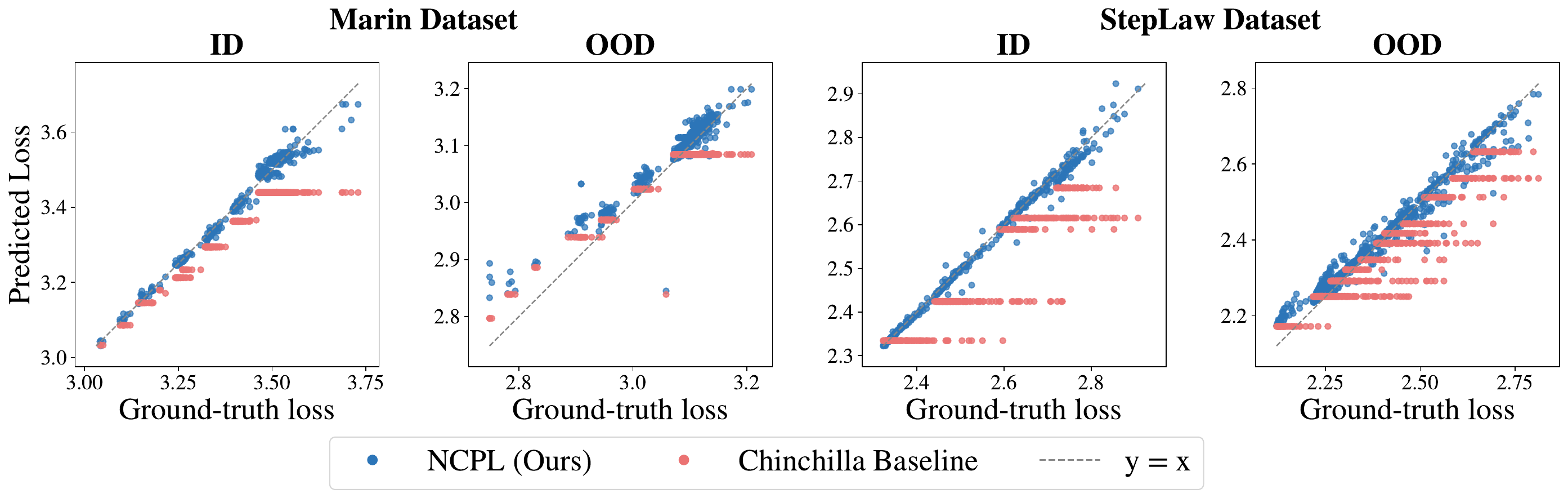}
    \caption{Predicted vs. ground-truth final pretraining loss of NCPL trained from scratch (1.7B model). }
    \label{fig:app_scatter_qwen}
\end{figure}

\begin{figure}[t]
    \centering
    \includegraphics[width=\linewidth]{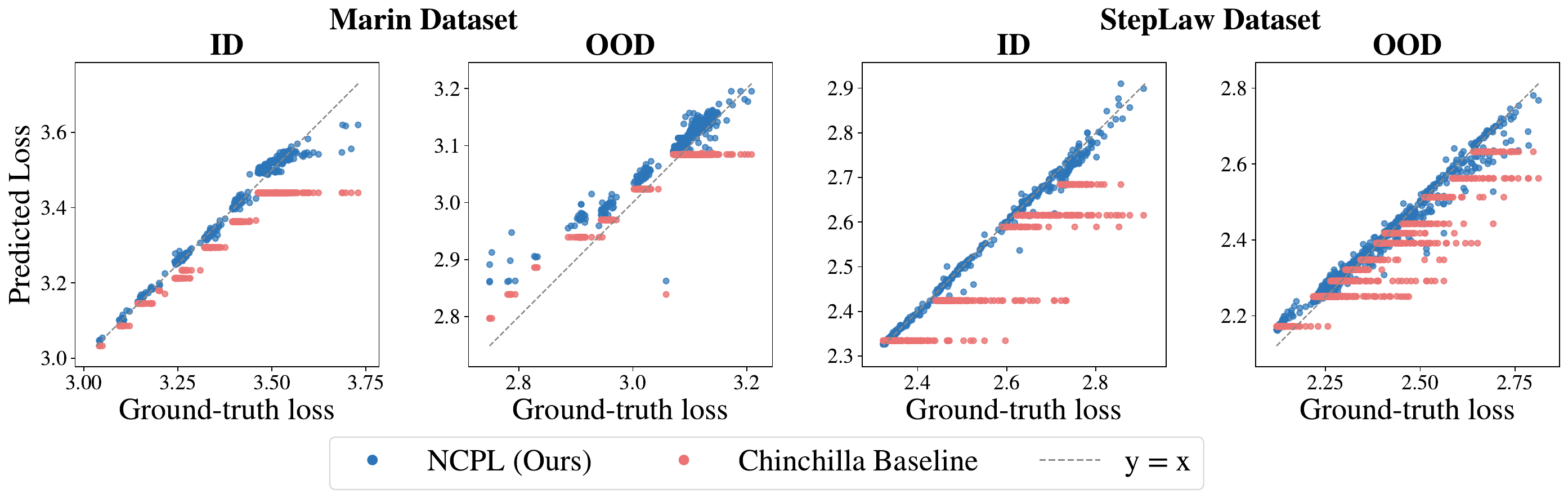}
    \caption{Predicted vs. ground-truth final pretraining loss of NCPL trained from scratch (135M model). }
    \label{fig:app_scatter_smol}
\end{figure}

Fig. \ref{fig:app_scatter_qwen} and \cref{fig:app_scatter_smol} show the predicted vs. ground-truth final pretraining loss of \predictor\ trained from scratch, using 1.7B model and 135M model respectively.

\begin{table*}[t]
\caption{Ablations on final-loss prediction on both ID and OOD splits. We compare NCPL (Ours) with two ablations: removing residual prediction and removing numerical tokens. We report mean absolute error (MAE), root mean squared error (RMSE), and Spearman correlation ($\rho$).}
\centering
\small
\setlength{\tabcolsep}{4pt}
\begin{tabular}{cccccccc}
\toprule
\multirow{3}{*}{\textbf{Data}}
& \multirow{3}{*}{\textbf{Method}}
& \multicolumn{6}{c}{\textbf{Final-loss prediction}} \\
\cmidrule(lr){3-8}
& & \multicolumn{3}{c}{In-distribution}
& \multicolumn{3}{c}{Out-of-distribution} \\
\cmidrule(lr){3-5}\cmidrule(lr){6-8}
& & MAE$\downarrow$ & RMSE$\downarrow$ & $\rho$ $\uparrow$
& MAE$\downarrow$ & RMSE$\downarrow$ & $\rho$ $\uparrow$ \\
\midrule

\multirow{3}{*}{Marin}
& \multicolumn{1}{c}{NCPL (Ours)}
& \textbf{0.0109} & \textbf{0.0169} & \textbf{0.9813}
& \textbf{0.0168} & \textbf{0.0239} & \textbf{0.9299} \\
& \multicolumn{1}{c}{No Residual Prediction}
& 0.0133 & 0.0195 & 0.9745
& 0.1503 & 0.1576 & 0.9264 \\
& \multicolumn{1}{c}{No Numerical Tokens}
& 0.0123 & 0.0243 & 0.9665
& 0.0217 & 0.0288 & 0.9267 \\
\midrule

\multirow{3}{*}{StepLaw}
& \multicolumn{1}{c}{NCPL (Ours)}
& \textbf{0.0097} & \textbf{0.0158} & 0.9948
& \textbf{0.0223} & \textbf{0.0345} & \textbf{0.9837} \\
& \multicolumn{1}{c}{No Residual Prediction}
& 0.0208 & 0.0301 & \textbf{0.9968}
& 0.0988 & 0.1134 & 0.9607 \\
& \multicolumn{1}{c}{No Numerical Tokens}
& 0.0123 & 0.0212 & 0.9906
& 0.0357 & 0.0428 & 0.9763 \\
\bottomrule
\end{tabular}
\vspace{-0.2in}
\label{tab:ablation-designchoice}
\end{table*}

